\documentclass[11pt,letterpaper]{article}
\usepackage[margin=1in,headheight=14pt]{geometry}
\usepackage[T1]{fontenc}
\usepackage{amsmath,amsthm}
\usepackage{newtxtext,newtxmath}
\usepackage{microtype,graphicx,xcolor}
\usepackage{booktabs,tabularx,array,longtable}
\usepackage[round,authoryear]{natbib}
\usepackage{tikz}
\usetikzlibrary{arrows.meta,positioning,calc,fit,backgrounds}
\usepackage{algorithm,algpseudocode,enumitem,titlesec,fancyhdr}
\definecolor{reportblue}{HTML}{1B5876}
\definecolor{computeblue}{HTML}{E3EFF8}
\definecolor{selectyellow}{HTML}{FFF2CE}
\definecolor{payloadpeach}{HTML}{FBE5D5}
\definecolor{bankgray}{HTML}{F2F3F4}
\definecolor{outputgreen}{HTML}{E2F1DD}
\usepackage[colorlinks=true,linkcolor=reportblue,citecolor=reportblue,urlcolor=reportblue]{hyperref}
\usepackage[capitalise,nameinlink]{cleveref}
\setlist[itemize]{leftmargin=1.5em,itemsep=1pt,topsep=3pt}
\setlist[enumerate]{leftmargin=1.8em,itemsep=1pt,topsep=3pt}
\titleformat{\section}{\large\bfseries}{\thesection.}{0.5em}{}
\titleformat{\subsection}{\normalsize\bfseries}{\thesubsection.}{0.5em}{}
\titleformat{\subsubsection}{\normalsize\bfseries}{\thesubsubsection.}{0.5em}{}
\titlespacing*{\section}{0pt}{14pt plus 2pt minus 1pt}{6pt}
\titlespacing*{\subsection}{0pt}{10pt plus 2pt minus 1pt}{4pt}
\titlespacing*{\subsubsection}{0pt}{7pt plus 1pt minus 1pt}{3pt}
\newtheorem{definition}{Definition}
\newtheorem{proposition}{Proposition}
\newtheorem{corollary}{Corollary}

\theoremstyle{remark}

\newcommand{\DT}{\textsc{DeepThink}}
\newcommand{\opnorm}[1]{\left\lVert#1\right\rVert_{\mathrm{op}}}
\newcolumntype{Y}{>{\raggedright\arraybackslash}X}
\algrenewcommand{\algorithmicrequire}{\textbf{Input:}}
\algrenewcommand{\algorithmicensure}{\textbf{Output:}}
\hypersetup{
  pdftitle={DeepThink: A Conceptual Model for Knowledge-Addressable Recurrent Language Modeling},
  pdfauthor={Chen Weiguo},
  pdfsubject={Architecture, static realizations, and pretrained migration},
  pdfkeywords={language models, knowledge units, recurrence, parameter assembly, pretrained migration}
}

\begin{document}
\begin{center}
{\fontsize{24}{28}\selectfont\bfseries DeepThink: A Conceptual Model for\\[3pt]
Knowledge-Addressable Recurrent\\[3pt]
Language Modeling}

\vspace{9pt}
{\large Chen Weiguo}\\[4pt]
{\normalsize NUDT}\\[3pt]
{\small\href{mailto:chenweiguo@nudt.edu.cn}{chenweiguo@nudt.edu.cn}}

\vspace{6pt}
{\small 3 November 2026}
\end{center}
\vspace{5pt}
\begin{abstract}
This manuscript presents DeepThink, a knowledge-addressable framework for recurrent
language modeling. A Knowledge Bank stores typed Knowledge Units, a Knowledge
Activator selects units from the current execution state, Knowledge Assembly
produces an executable configuration, and a Shared Thinking Operator updates
that state. An output stage then organizes generation. This organization
exposes knowledge capacity, activation width, and recurrent depth as separate
design variables, supporting both complete parameter packages and fine-grained
expert or low-rank modules. We formalize the interfaces and their execution
semantics, including state-dependent selection, parameter-space assembly,
and thinking followed by generation. For compatible templates and complete
runtime-state configurations, we establish an exact Fixed-Schedule Realization
of an existing fixed-depth model. Exact shared-base reparameterization and
pretrained output-tail reuse provide complementary initialization
constructions. Together, these results define a route from pretrained models
to dynamically selected and assembled recurrent computation. A synthesis of
conditional computation, recurrent models, modular parameters, and knowledge
interfaces positions DeepThink within the research landscape and identifies
the design questions arising from their integration.

\end{abstract}
\vspace{4pt}

\clearpage
\tableofcontents
\clearpage
\label{report:main-start}

\section{Overview and Architectural Motivation}
\label{sec:overview}

\DT{} organizes language modeling around addressable parameter content and
reusable state updates. A model selects a knowledge configuration, applies
a shared computational template, and uses the resulting state to guide the
next invocation. The framework makes knowledge storage, execution order,
and output organization explicit, providing both a vocabulary for dynamic
architectures and a route for reusing pretrained models.
\Cref{sec:landscape} develops its connections to sparse experts, recurrence,
latent refinement, and modular parameters.

\subsection{Knowledge Content and Repeated Computation}
\subsubsection{Computation and Memory in the von Neumann Perspective}
The von Neumann architecture provides a useful starting analogy for separating
stored content from its execution. Its classical organization distinguishes
arithmetic and control from memory, with memory holding both instructions
and data \citep{vonneumann1945}. Computation proceeds by accessing stored
information through an execution interface. The associated data-transfer
constraints were later discussed as the von Neumann bottleneck
\citep{backus1978}.

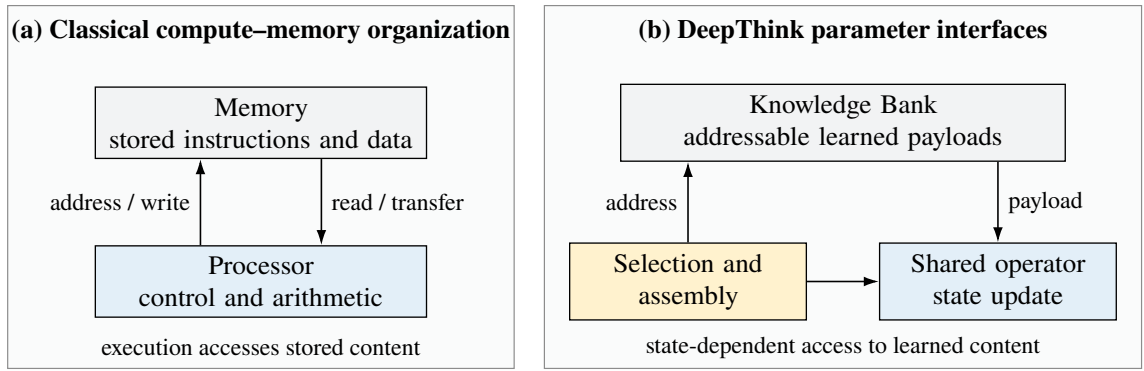
\begin{figure}[htbp]
\centering
\begin{tikzpicture}[
  x=1cm,y=1cm,font=\small,
  block/.style={draw=black,line width=0.5pt,rectangle,align=center,
    minimum height=0.85cm,inner sep=4pt},
  flow/.style={-{Latex[length=2mm,width=1.3mm]},line width=0.6pt},
  annotation/.style={font=\footnotesize,align=center}
]
  \begin{scope}[on background layer]
    \draw[draw=black!45,fill=bankgray!35,line width=0.45pt]
      (0,0) rectangle (6.7,4.8);
    \draw[draw=black!45,fill=bankgray!35,line width=0.45pt]
      (7.1,0) rectangle (15.0,4.8);
  \end{scope}
  \node[font=\small\bfseries] at (3.35,4.45)
    {(a) Classical compute--memory organization};
  \node[font=\small\bfseries] at (11.05,4.45)
    {(b) DeepThink parameter interfaces};
  \node[block,fill=bankgray,text width=4.1cm] (memory) at (3.35,3.25)
    {Memory\\stored instructions and data};
  \node[block,fill=computeblue,text width=4.1cm] (processor) at (3.35,1.15)
    {Processor\\control and arithmetic};
  \draw[flow] ([xshift=-8mm]processor.north) --
    node[annotation,left] {address / write} ([xshift=-8mm]memory.south);
  \draw[flow] ([xshift=8mm]memory.south) --
    node[annotation,right] {read / transfer} ([xshift=8mm]processor.north);
  \node[annotation] at (3.35,0.28) {execution accesses stored content};
  \node[block,fill=bankgray,text width=5.6cm] (bank) at (11.05,3.25)
    {Knowledge Bank\\addressable learned payloads};
  \node[block,fill=selectyellow,text width=2.85cm] (control) at (9.0,1.15)
    {Selection and\\assembly};
  \node[block,fill=computeblue,text width=2.85cm] (operator) at (13.1,1.15)
    {Shared operator\\state update};
  \draw[flow] (control.north) --
    node[annotation,left] {address} ([xshift=-20.5mm]bank.south);
  \draw[flow] ([xshift=20.5mm]bank.south) --
    node[annotation,right] {payload} (operator.north);
  \draw[flow] (control.east) -- (operator.west);
  \node[annotation] at (11.05,0.28) {state-dependent access to learned content};
\end{tikzpicture}
\caption{Compute--memory separation as an architectural analogy.
Panel (a) depicts the classical distinction between a processor and a memory
containing instructions and data. Panel (b) transfers the organizational idea
to DeepThink: learned payloads are stored in a bank and addressed through
selection and assembly to configure a shared state-update operator.
The drawing concerns interface roles and does not specify a hardware memory
layout. Input/output peripherals and recurrent feedback are omitted.}
\label{fig:compute-memory}
\end{figure}

DeepThink applies this organizational idea to learned parameter content.
A Knowledge Bank makes configurations addressable, while selection and
assembly supply a reusable operator with the content needed for each update.
\Cref{fig:compute-memory} illustrates the relationship. Knowledge Units are
learned computational payloads rather than machine instructions or symbolic
facts. Their separation is an architectural interface choice: bank access,
parameter transfer, and state updates still require physical computation
and memory traffic.

\suppressfloats[t]

\subsubsection{Parameterized Content in Language Models}
Learned parameters provide the content used by a language model and shape the
transformations applied to that content. These functions are distributed
across the network. Analyses of Transformer feedforward layers as key-value
memories provide one concrete account of parameterized content
\citep{ffmemory2020}. DeepThink introduces addressable content configurations
and reusable rules for coordinating their application. The motivating
question is how knowledge can be selected and recombined across a sequence
of state updates while retaining a stable computational interface.

\subsubsection{Repeated Templates and Depth-Specific Parameters}
Many backbones repeat a computational template with depth-specific parameters,
as in the repeated attention and feedforward blocks of the Transformer
\citep{vaswani2017}. Let $x$ denote the model input, $\mathcal C$ the permitted
conditioning and execution context, and $D\ge1$ the number of backbone
transitions. The superscript $\ell\in\{0,\ldots,D\}$ indexes backbone depth.
The complete state $X^{(\ell)}=(H^{(\ell)},M^{(\ell)})$ contains hidden
representations $H^{(\ell)}$ and persistent runtime state $M^{(\ell)}$.
Let $E_*$ be the pretrained initialization map, $F_\ell$ the transition at
depth $\ell$, $\omega_\ell$ its complete configuration, and $G_*$ the original
output map. The subscript $*$ marks pretrained quantities. Then
\begin{equation}
 \begin{aligned}
 X^{(0)}&=E_*(x;\mathcal C),\\
 X^{(\ell)}&=F_\ell(X^{(\ell-1)};\omega_\ell,\mathcal C),
 \quad \ell=1,\ldots,D,\\
 Y_*&=G_*(X^{(D)};\mathcal C),
 \end{aligned}
 \label{eq:original-backbone}
\end{equation}
This repeated-template structure suggests separating the execution template
from its depth-specific configurations. Each configuration retains the
parameters that give an invocation its particular behavior. For heterogeneous
backbones, the same organization applies within module types: a family of
typed operators preserves the distinctions in the original computation graph.

\subsection{The DeepThink Perspective}
\subsubsection{Addressable Knowledge and Shared State Updates}
A Knowledge Bank stores numbered Knowledge Units. A Knowledge Activator selects
units using the current state; Knowledge Assembly converts the selection into
a configuration for a Shared Thinking Operator. The updated state drives the
next selection. A stable identifier can be called at different steps, and
the same step can invoke different configurations for different inputs. This
exposes two choices: which content is called and when it is called. A fixed
schedule provides an initial policy for these choices.

\subsubsection{Knowledge Capacity, Activation Width, and Recurrent Depth}
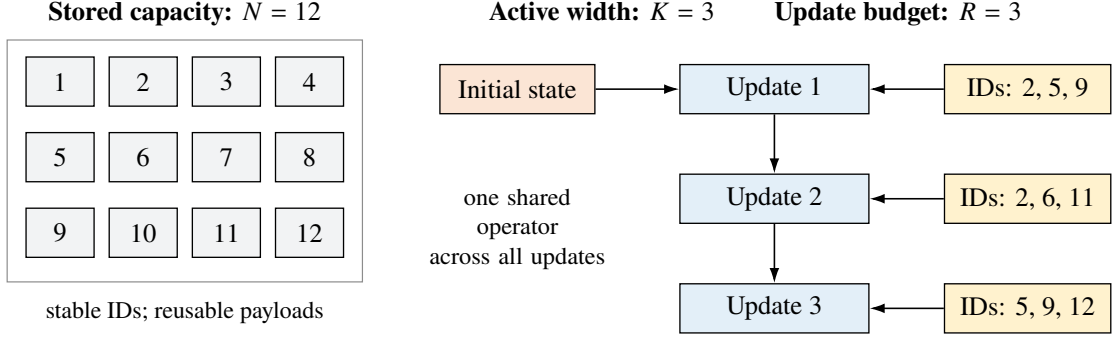
\begin{figure}[htbp]
\centering
\begin{tikzpicture}[
  x=1cm,y=1cm,font=\small,
  block/.style={draw=black,line width=0.5pt,rectangle,align=center,
    minimum height=0.65cm,inner sep=4pt},
  flow/.style={-{Latex[length=2mm,width=1.3mm]},line width=0.6pt},
  annotation/.style={font=\footnotesize,align=center}
]
  \node[font=\small\bfseries] at (2.45,4.45) {Stored capacity: $N=12$};
  \foreach \row in {0,1,2} {
    \foreach \col in {0,1,2,3} {
      \pgfmathtruncatemacro{\unitid}{4*\row+\col+1}
      \node[block,fill=bankgray,minimum width=0.9cm]
        at (0.8+1.1*\col,3.55-1.0*\row) {\unitid};
    }
  }
  \draw[draw=black!45,line width=0.45pt] (0.1,0.9) rectangle (4.8,4.1);
  \node[annotation] at (2.45,0.5) {stable IDs; reusable payloads};
  \node[font=\small\bfseries] at (10.0,4.45)
    {Active width: $K=3$ \qquad Update budget: $R=3$};
  \node[block,fill=payloadpeach,minimum width=2.05cm] (start) at (6.85,3.45)
    {Initial state};
  \node[block,fill=computeblue,minimum width=2.5cm] (first) at (10.25,3.45)
    {Update 1};
  \node[block,fill=computeblue,minimum width=2.5cm] (second) at (10.25,2.0)
    {Update 2};
  \node[block,fill=computeblue,minimum width=2.5cm] (third) at (10.25,0.55)
    {Update 3};
  \node[block,fill=selectyellow,minimum width=2.2cm] (sel1) at (13.6,3.45)
    {IDs: 2, 5, 9};
  \node[block,fill=selectyellow,minimum width=2.2cm] (sel2) at (13.6,2.0)
    {IDs: 2, 6, 11};
  \node[block,fill=selectyellow,minimum width=2.2cm] (sel3) at (13.6,0.55)
    {IDs: 5, 9, 12};
  \draw[flow] (start.east) -- (first.west);
  \draw[flow] (first.south) -- (second.north);
  \draw[flow] (second.south) -- (third.north);
  \draw[flow] (sel1.west) -- (first.east);
  \draw[flow] (sel2.west) -- (second.east);
  \draw[flow] (sel3.west) -- (third.east);
  \node[annotation,text width=2.5cm] at (6.85,1.6)
    {one shared operator\\across all updates};
\end{tikzpicture}
\caption{Independent controls over stored content, active content, and
recurrent updates. This illustrative schedule stores 12 units, selects three
at each decision, and executes three updates. Units can recur across steps,
and the selected set can change while the operator is shared. Increasing the
bank size does not prescribe the active width or update budget. The example
describes configuration choices rather than measured performance.}
\label{fig:capacity-controls}
\end{figure}

The bank size $N$, activation width $K$, and recurrence budget $R$ are separate
positive-integer configuration variables, with $1\le K\le N$ and $R\ge1$. They control available
units, units selected per routing decision, and state updates. Increasing $N$
need not increase $K$ or $R$, although routing cost can still depend on $N$.
The content and size of each unit determine the capacity represented by that count.
\Cref{fig:capacity-controls} makes these three controls concrete. Sparse
expert models already separate stored capacity from active expert use
\citep{shazeer2017,switch2021}; shared-depth models separately expose repeated
computation \citep{ut2018}. DeepThink specifies both through one execution
interface.

\suppressfloats[t]

\subsection{Organizing Knowledge and Computation}
\subsubsection{Interface Roles and Parameter Ownership}
DeepThink gives storage, selection, assembly, state transition, and output
distinct roles. The bank owns content payloads, the activator owns invocation
decisions, the assembly interface specifies composition, and the operator
coordinates state updates. This separation makes parameter configurations
independently addressable while allowing shared parameters to participate in
the execution. An implementation can fuse these operations for efficiency
while preserving their interfaces. \Cref{prop:fixed} connects this organization
to the computation of an existing model.

\subsubsection{Reusable Content and Shared Thinking}
The central intuition is that content can be reused across different
computational situations while a common operator organizes its application.
The state update changes what the model needs next: a configuration useful
for interpreting an input may give way to one useful for integrating
information or preparing an output. State-conditioned activation makes this
change explicit. Learning objectives can encourage Knowledge Units to supply
reusable contributions and the Shared Thinking Operator to coordinate those
contributions across successive updates. The terms \emph{knowledge} and
\emph{thinking} describe these intended computational roles.

\section{Core Architecture and Execution}
\label{sec:core}

\begin{figure}[htbp]
\centering
\begin{tikzpicture}[
  x=1cm,y=1cm,font=\small,
  block/.style={draw=black,line width=0.5pt,rectangle,align=center,
    minimum height=0.8cm,inner sep=4pt},
  flow/.style={-{Latex[length=2mm,width=1.3mm]},line width=0.6pt},
  annotation/.style={font=\footnotesize,align=center}
]
  \node[block,fill=bankgray,minimum width=3.6cm] (bank) at (7.1,6.0)
    {Knowledge Bank $\mathcal K$\\
     $\kappa_1,\ \kappa_2,\ \ldots,\ \kappa_N$};
  \node[block,fill=selectyellow,minimum width=2.4cm] (activate) at (4.2,4.2)
    {Knowledge\\Activator $A_\psi$};
  \node[block,fill=selectyellow,minimum width=2.1cm] (select) at (7.1,4.2)
    {Top-$K$\\select $K$ of $N$};
  \node[block,fill=selectyellow,minimum width=2.4cm] (assemble) at (10.0,4.2)
    {Knowledge\\Assembly $\Gamma$};
  \node[block,fill=bankgray,minimum width=2.4cm] (state) at (4.2,1.7)
    {Current state\\$X_t=(H_t,M_t)$};
  \node[block,fill=computeblue,minimum width=2.4cm] (think) at (10.0,1.7)
    {Shared Thinking\\Operator $\Phi_\theta$};
  \node[block,fill=bankgray,minimum width=2.0cm] (next) at (13.0,1.7)
    {Updated state\\$X_{t+1}$};
  \node[block,fill=outputgreen,minimum width=2.2cm] (output) at (13.0,4.2)
    {Output module\\$G_\varphi$};
  \node[block,fill=payloadpeach,minimum width=2.7cm] (initialize) at (4.2,-0.65)
    {Input Representation\\State Initialization};

  \begin{scope}[on background layer]
    \draw[draw=black!45,fill=bankgray!55,line width=0.45pt]
      (2.85,0.65) rectangle (11.35,5.0);
  \end{scope}
  \node[anchor=north west,font=\small\bfseries] at (3.0,5.35)
    {Recurrent module};

  \draw[flow] (initialize.north) -- (state.south);
  \draw[flow] (state.north) -- node[annotation,left] {$q_t$} (activate.south);
  \draw[flow] (activate.east) -- node[annotation,above] {$s_t$} (select.west);
  \draw[flow] (bank.south) -- (select.north);
  \draw[flow] (select.east) -- node[annotation,below=18pt] {$S_t,\alpha_t$} (assemble.west);
  \draw[flow] (assemble.south) -- node[annotation,left] {$\widetilde\kappa_t$} (think.north);
  \draw[flow] (state.east) -- (think.west);
  \draw[flow] (think.east) -- (next.west);
  \draw[flow] (next.north) -- node[annotation,right] {at stopping} (output.south);
  \draw[flow] (next.south) -- (13.0,0.25) -- (4.2,0.25) -- (state.south);
  \fill (4.2,0.25) circle (1.1pt);
  \node[annotation,anchor=north] at (8.55,0.14)
    {updated state returns for the next step};
\end{tikzpicture}
\caption{The DeepThink execution interface. The current complete state forms a
query for the Knowledge Activator; Top-$K$ selection addresses the external
Knowledge Bank, and Knowledge Assembly supplies the current configuration to
one Shared Thinking Operator. The updated state feeds back into the next
iteration. The output module receives the final state after $R_{\mathrm{exec}}$ iterations,
equal to $R$ under a fixed budget.
The query interface and external execution context are omitted for clarity.}
\label{fig:core}
\end{figure}
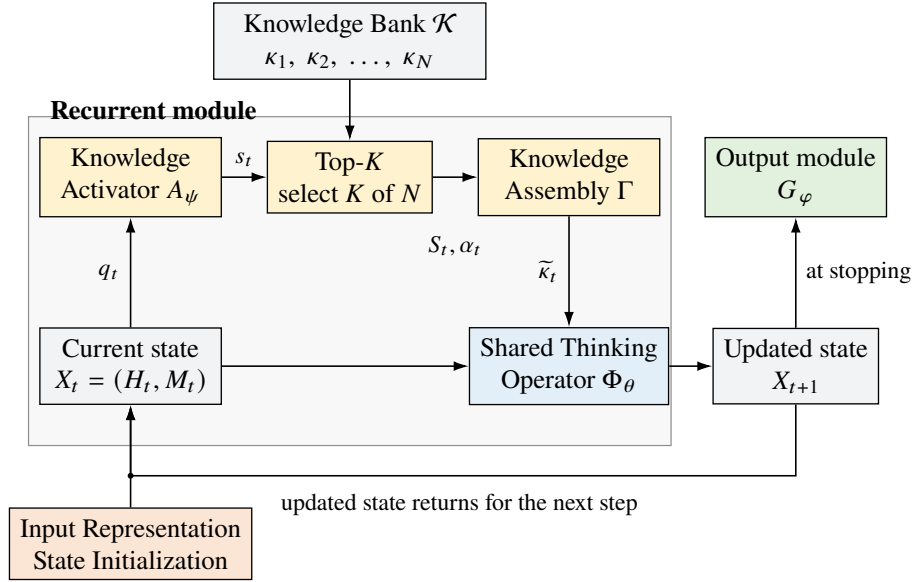

The following interfaces define a family of systems. Attention,
normalization, nonlinearities, position processing, and dependency graphs
belong to the chosen execution template. Its configuration specifies their
behavior and the state passed between invocations.
\Cref{fig:core} summarizes the invocation and feedback paths.

\subsection{Notation and Execution State}
\subsubsection{Hidden Representations and Complete Runtime State}
The nonnegative integer $t$ indexes recurrent updates, independently of
backbone depth $\ell$. Write $X_t=(H_t,M_t)$ for the complete state before
update $t$. $H_t$ contains hidden representations, and $M_t$ includes
required workspace, caches, counters, and other persistent state.
The parameter groups $\xi$, $\psi$, $\theta$, and $\varphi$ belong to the
initialization map, activator, shared operator, and output stage, respectively;
any trainable query and stopping-control parameters are included in $\psi$.
A group can be empty when its interface has no parameters. Initialization is
$X_0=E_\xi(x;\mathcal C)$, where $E_\xi$ constructs the complete input state. Carrying the complete state makes initialization,
recurrence, and output-stage handoff part of one execution specification.
For input length $n\ge1$ and hidden width $d\ge1$, a common representation is
$H_t\in\mathbb R^{n\times d}$;
incremental inference can instead expose one position with previously stored
state. These two execution modes must agree on the information available to
each invocation if equivalent autoregressive behavior is intended.

\subsubsection{External Context and Autoregressive Boundaries}
The context $\mathcal C$ supplies legal conditioning, masks, positions, and
execution metadata. The positive integer $p$ indexes target-token positions.
At position $p$, activation and assembly may depend
only on the available prefix and permitted external information. A sequence
summary containing future target tokens is inadmissible. The recurrence index
$t$ and token index $p$ describe different axes of computation.

\subsection{Knowledge Units and Typed Banks}
\subsubsection{Identifiers, Parameter Payloads, and Compatible Interfaces}
\begin{definition}[Typed Knowledge Unit]
\label[definition]{def:unit}
A Knowledge Unit is $\kappa_i=(i,\rho_i,\vartheta_i)$, where $i$ is a stable identifier,
$\rho_i$ specifies its execution interface, and $\vartheta_i$ is a parameter payload or
functional module compatible with that interface. A Knowledge Bank is
$\mathcal K=\{\kappa_i\}_{i=1}^{N}$. Units can be assembled at an interface only
when their types agree, or an explicit adapter supplies compatibility.
\end{definition}
Types specify dimensions and calling semantics. A complete payload can
contain a parameter dictionary and configuration metadata. It may also
encapsulate a functional module or its structural constants. The typed
interface determines which components can be combined numerically and which
are selected as a whole.

\subsubsection{Addressable Knowledge Modules}
Identifiers support selection, replacement, freezing, and versioning.
An expert network, an adapter, and a complete block configuration are
alternative payload forms. Units can encode overlapping content or
participate in coordinated computations. Their execution interfaces specify
how those contributions enter the state update.

\subsection{Knowledge Activation}
\subsubsection{State-Conditioned Selection and Top-K Activation}
Let $\mathcal I_t\subseteq\{1,\ldots,N\}$ be the identifiers eligible at
update $t$ under the current type and execution constraints. A query map $Q$
constructs $q_t\in\mathbb R^{d_q}$, where $d_q\ge1$ is the query width.
The Knowledge Activator $A_\psi$ returns real scores
$s_t=(s_{t,i})_{i\in\mathcal I_t}$ over those eligible identifiers:
\begin{equation}
 q_t=Q(X_t,\mathcal C,t),\qquad
 s_t=A_\psi(q_t,X_t,\mathcal C,t;\mathcal I_t),\qquad
 S_t=\operatorname{TopK}_K(s_t;\mathcal I_t).
 \label{eq:activation}
\end{equation}
Here $\operatorname{TopK}_K$ returns the $K$ highest-scoring compatible
identifiers, and $S_t\subseteq\mathcal I_t$ is their support set.
For selected identifier $i$, one optional coefficient rule is
\begin{equation}
 \alpha_{t,i}=
 \frac{\exp(s_{t,i})}{\sum_{u\in S_t}\exp(s_{t,u})},
 \qquad i\in S_t.
 \label{eq:selected-weights}
\end{equation}
Selection and weighting are separate policies; normalization is not a
requirement of every assembly rule.
The coefficient vector is $\alpha_t=(\alpha_{t,i})_{i\in S_t}$, and $u$ in
the denominator indexes selected identifiers.
Each invocation requires \(1\le K\le|\mathcal I_t|\), where $|\cdot|$ denotes
set cardinality;
scores and Top-\(K\) selection are restricted to that eligible bank.
Tie resolution belongs to the selection policy. A state-dependent policy
can retain a selection across several iterations or change it as the
representation evolves. Even with the same selected units, different inputs
produce different state transitions.

\subsubsection{Fixed Policies and Routing Granularity}
A fixed policy may choose units by execution position instead of input content.
This policy supports a Fixed-Schedule Realization of an existing model.
Dynamic decisions may operate per prefix, token, sequence, or local interface.
Thus $K$ counts units per decision; different tokens can collectively access
more than $K$ distinct units in one sequence.
A fixed schedule can coexist with dynamic routing inside a stored payload,
as in an already trained expert block. ``Fixed'' then describes the outer
invocation sequence, not every decision within the original computation.
Maintaining this distinction is necessary when reusing sparse backbones.

\subsection{Knowledge Assembly}
\subsubsection{Parameter-Space and Function-Space Composition}
\begin{figure}[htbp]
\centering
\begin{tikzpicture}[
  x=1cm,y=1cm,font=\small,
  block/.style={draw=black,line width=0.5pt,rectangle,align=center,
    minimum height=0.8cm,inner sep=4pt},
  flow/.style={-{Latex[length=2mm,width=1.3mm]},line width=0.6pt},
  annotation/.style={font=\footnotesize,align=center}
]
  \begin{scope}[on background layer]
    \draw[draw=black!45,fill=bankgray!35,line width=0.45pt]
      (0,0) rectangle (6.9,6.1);
    \draw[draw=black!45,fill=bankgray!35,line width=0.45pt]
      (7.35,0) rectangle (15.0,6.1);
  \end{scope}
  \node[font=\small\bfseries] at (3.45,5.75)
    {(a) Combine parameters, then evaluate};
  \node[font=\small\bfseries] at (11.18,5.75)
    {(b) Evaluate modules, then combine};
  \node[block,fill=payloadpeach,text width=2.3cm] (w1) at (1.8,4.75) {$W_1$};
  \node[block,fill=payloadpeach,text width=2.3cm] (w2) at (5.1,4.75) {$W_2$};
  \node[block,fill=selectyellow,text width=4.5cm] (sum) at (3.45,3.25)
    {$\sum_{i\in S_t}\alpha_{t,i}W_i$};
  \node[block,fill=computeblue,text width=4.5cm] (f) at (3.45,1.85)
    {One nonlinear module $f$};
  \node[block,fill=outputgreen,text width=4.5cm,minimum height=0.6cm]
    (outa) at (3.45,0.65) {Scalar example: output $0$};
  \draw[flow] (w1.south) -- ([xshift=-16.5mm]sum.north);
  \draw[flow] (w2.south) -- ([xshift=16.5mm]sum.north);
  \draw[flow] (sum.south) -- (f.north);
  \draw[flow] (f.south) -- (outa.north);
  \node[annotation] (za) at (0.5,1.85) {$z$};
  \draw[flow] (za.east) -- (f.west);
  \node[block,fill=bankgray,text width=4.6cm] (zb) at (11.18,4.75) {Input $z$};
  \node[block,fill=computeblue,text width=2.75cm] (f1) at (9.25,3.25)
    {$f(z;W_1)$\\example: $1$};
  \node[block,fill=computeblue,text width=2.75cm] (f2) at (13.1,3.25)
    {$f(z;W_2)$\\example: $0$};
  \node[block,fill=selectyellow,text width=5.5cm] (merge) at (11.18,1.85)
    {$\sum_{i\in S_t}\alpha_{t,i} f(z;W_i)$};
  \node[block,fill=outputgreen,text width=5.5cm,minimum height=0.6cm]
    (outb) at (11.18,0.65) {Scalar example: output $1/2$};
  \draw[flow] ([xshift=-19.25mm]zb.south) -- (f1.north);
  \draw[flow] ([xshift=19.25mm]zb.south) -- (f2.north);
  \draw[flow] (f1.south) -- ([xshift=-19.25mm]merge.north);
  \draw[flow] (f2.south) -- ([xshift=19.25mm]merge.north);
  \draw[flow] (merge.south) -- (outb.north);
\end{tikzpicture}
\caption{Assembly semantics change the computed function. Panel (a) combines
compatible matrices before applying a nonlinear module; panel (b) combines
the independently evaluated module outputs. For the scalar ReLU example
defined in the text, the outputs are $0$ and $1/2$, respectively, despite
identical inputs, parameters, and coefficients. The distinction disappears
for the purely linear increments in \cref{eq:assembled-linear}.}
\label{fig:assembly-semantics}
\end{figure}
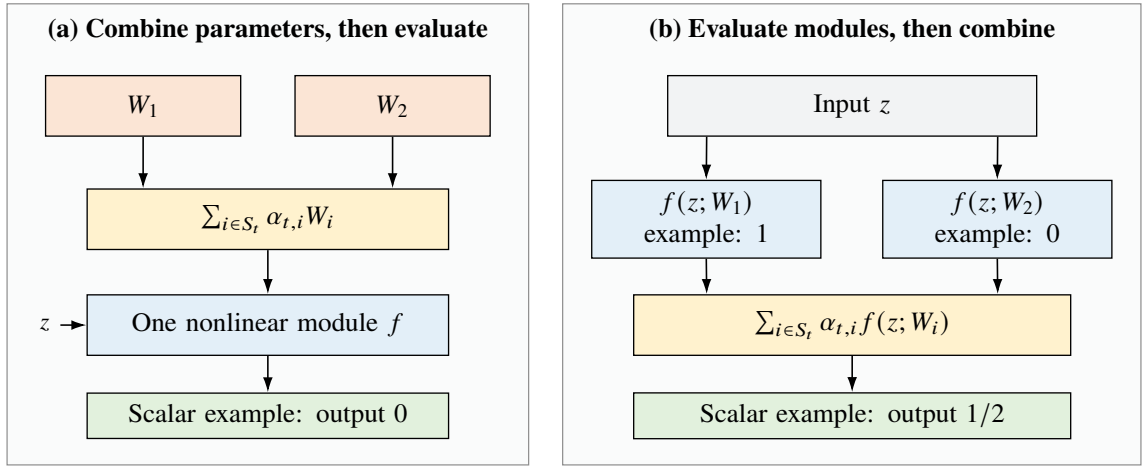

\begin{definition}[Knowledge Assembly]
\label[definition]{def:assembly}
Knowledge Assembly is a typed map
$\widetilde\kappa_t=\Gamma(\mathcal K|_{S_t},\alpha_t)$ from selected payloads
and optional coefficients to a configuration accepted by an execution
interface. Here $\mathcal K|_{S_t}=\{\kappa_i:i\in S_t\}$ is the selected
bank restriction, $\Gamma$ is the assembly map, and
$\widetilde\kappa_t$ is its executable configuration.
Its contract specifies parameter combination, function-output
combination, or sequential composition, together with any required order.
\end{definition}
At a matrix-parameterized interface, let $z\in\mathbb R^{d_{\mathrm{in}}}$
be a row-vector input and $W_i\in\mathbb R^{d_{\mathrm{in}}\times d_{\mathrm{out}}}$
compatible unit matrices. The positive integers $d_{\mathrm{in}}$ and
$d_{\mathrm{out}}$ are the input and output widths. Let $f(z;W)$ be the module
output in $\mathbb R^{d_{\mathrm{out}}}$, with other settings held fixed.
For nonlinear $f$, parameter mixing and output mixing generally differ:
\begin{equation}
 f\!\left(z;\textstyle\sum_{i\in S_t}\alpha_{t,i}W_i\right)
 \ne \sum_{i\in S_t}\alpha_{t,i} f(z;W_i).
 \label{eq:assembly-nonequivalence}
\end{equation}
\Cref{fig:assembly-semantics} illustrates the distinction. For a scalar $a$,
define $\operatorname{ReLU}(a)=\max(0,a)$. With $S_t=\{1,2\}$,
$z=1$, $W_1=1$, $W_2=-1$, and $\alpha_{t,1}=\alpha_{t,2}=1/2$,
the module $f(z;W)=\operatorname{ReLU}(zW)$ yields zero under parameter
mixing and $1/2$ under output mixing. Parameter merging and sparse
expert-output mixtures have corresponding precedents in Soft Merging of
Experts with Adaptive Routing (SMEAR) and sparse mixture-of-experts (MoE) models
\citep{smear2023,shazeer2017}.

\suppressfloats[t]

Linear increments can be
evaluated as selected branches implementing the effective parameter sum;
nonlinear experts follow the specified parameter-space or function-space
composition. A coarse configuration can bind several local interfaces
jointly. Assembly can be implemented through direct branch evaluation or
parameter materialization.

\subsubsection{Sets, Ordered Selections, and Assembly Semantics}
The notation $S_t$ suffices for permutation-invariant combinations. Ordered
composition instead uses a list $I_t=(i_{t,1},\ldots,i_{t,K})$, with an explicit
order policy. Ranking, execution order, and coefficient indexing must agree.
A singleton-preserving contract returns the original unit's execution
configuration when its sole coefficient is one.

\subsection{Shared Thinking and Output}
\subsubsection{Recurrent State Transitions and Stopping}
\begin{definition}[DeepThink Execution]
\label[definition]{def:execution}
Given initialization, activation, assembly, and a compatible Shared Thinking
Operator $\Phi_\theta$, execution applies
\begin{equation}
 X_{t+1}=\Phi_\theta(X_t;\widetilde\kappa_t,\mathcal C),
 \qquad t=0,\ldots,R_{\mathrm{exec}}-1,
 \label{eq:transition}
\end{equation}
where $R_{\mathrm{exec}}$ is the integer number of updates actually executed,
and $R_{\mathrm{exec}}=R$ for fixed-budget execution. An optional controller determines a
stopping time $1\le R_{\mathrm{exec}}\le R$ from legally available states. The output is
$Y=G_\varphi(X_{R_{\mathrm{exec}}};\mathcal C)$, where $G_\varphi$ is the
declared output-stage map.
\end{definition}
The shared operator combines reusable code and, optionally, shared trainable
parameters. Typed systems share operators within compatible interface classes.
Stopping decides how many updates are performed, whereas activation decides
which units contribute to an update. These policies can share observations
but remain separate choices. The recurrence budget exposes the number of
state updates as a separate control variable.

\suppressfloats[t]

\subsubsection{Output Interfaces and Execution Traces}
$Y$ may denote output logits, a distribution, or a generated object under a
specified decoding policy. An optional trace
$\tau=((X_t,S_t,\alpha_t))_{t=0}^{R_{\mathrm{exec}}-1}$ records state and invocation decisions;
the terminal state is $X_{R_{\mathrm{exec}}}$. These traces expose invocation paths, selection frequencies, and state
evolution for analysis. Generation-state updates
must follow the declared output contract.

\section{Parameterizations and Architectural Configurations}
\label{sec:configurations}

\subsection{Representing Knowledge Payloads}
\subsubsection{Complete Parameter Packages and Expert Modules}
A coarse unit may bind all parameters required by one block, including biases,
normalization parameters, and execution-specific settings. Alternatively, a
unit may be an expert function at a local interface. Complete packages are
particularly useful for preserving pretrained behavior; expert functions
require a declared combination rule and compatible intermediate states.

\subsubsection{Shared Bases and Low-Rank Increments}
At a linear interface, retain the row-vector convention above and choose a
shared matrix $W_{\mathrm{base}}\in\mathbb R^{d_{\mathrm{in}}\times d_{\mathrm{out}}}$.
Unit $i$ may store an increment $\Delta W_i=U_iV_i$ with positive integer
rank budget $r_i$. Its factors satisfy
$U_i\in\mathbb R^{d_{\mathrm{in}}\times r_i}$ and
$V_i\in\mathbb R^{r_i\times d_{\mathrm{out}}}$.
Let $b_{\mathrm{base}}\in\mathbb R^{d_{\mathrm{out}}}$ be a shared bias and
$\Delta b_i\in\mathbb R^{d_{\mathrm{out}}}$ an optional bias increment.
The effective matrix $\widetilde W_t$, effective bias $\widetilde b_t$, and
row-vector output $o_t\in\mathbb R^{d_{\mathrm{out}}}$ are
\begin{equation}
 \begin{aligned}
 \widetilde W_t&=W_{\mathrm{base}}+
 \sum_{i\in S_t}\alpha_{t,i}U_iV_i,\\
 \widetilde b_t&=b_{\mathrm{base}}+
 \sum_{i\in S_t}\alpha_{t,i}\Delta b_i,\\
 o_t&=z\widetilde W_t+\widetilde b_t
 =zW_{\mathrm{base}}+
 \sum_{i\in S_t}\alpha_{t,i}(zU_i)V_i+\widetilde b_t.
 \end{aligned}
 \label{eq:assembled-linear}
\end{equation}
Setting $\Delta b_i=0$ retains a fixed shared bias. Evaluating the selected
linear branches gives the same mathematical output as materializing
$\widetilde W_t$. Low-Rank Adaptation (LoRA) supplies a precedent for this
parameterization \citep{lora2021}. The rank budget controls the space
of representable increments; \Cref{sec:migration} develops exact and
approximate pretrained constructions.

\subsection{Thinking Followed by Generation}
\subsubsection{A Recurrent Core with an Output-Organizing Stage}
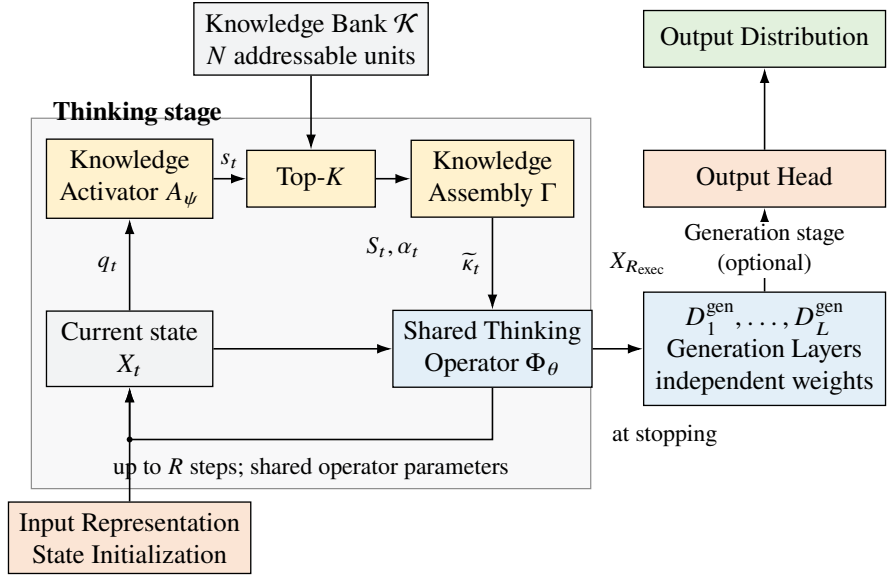
\begin{figure}[htbp]
\centering
\begin{tikzpicture}[
  x=1cm,y=1cm,font=\small,
  block/.style={draw=black,line width=0.5pt,rectangle,align=center,
    minimum height=0.75cm,inner sep=4pt},
  flow/.style={-{Latex[length=2mm,width=1.3mm]},line width=0.6pt},
  annotation/.style={font=\footnotesize,align=center}
]
  \node[block,fill=bankgray,minimum width=3.1cm] (bank) at (5.8,5.35)
    {Knowledge Bank $\mathcal K$\\$N$ addressable units};
  \node[block,fill=bankgray,minimum width=2.2cm] (state) at (3.4,1.25)
    {Current state\\$X_t$};
  \node[block,fill=selectyellow,minimum width=2.2cm] (activate) at (3.4,3.5)
    {Knowledge\\Activator $A_\psi$};
  \node[block,fill=selectyellow,minimum width=1.7cm] (select) at (5.8,3.5)
    {Top-$K$};
  \node[block,fill=selectyellow,minimum width=2.15cm] (assemble) at (8.2,3.5)
    {Knowledge\\Assembly $\Gamma$};
  \node[block,fill=computeblue,minimum width=2.2cm] (think) at (8.2,1.25)
    {Shared Thinking\\Operator $\Phi_\theta$};
  \node[block,fill=payloadpeach,minimum width=2.7cm] (input) at (3.4,-1.25)
    {Input Representation\\State Initialization};
  \node[block,fill=computeblue,minimum width=3.2cm,minimum height=1.1cm] (generate)
    at (11.8,1.25)
    {$D^{\mathrm{gen}}_1,\ldots,D^{\mathrm{gen}}_L$\\
     Generation Layers\\independent weights};
  \node[block,fill=payloadpeach,minimum width=3.2cm] (head) at (11.8,3.5)
    {Output Head};
  \node[block,fill=outputgreen,minimum width=3.2cm] (output) at (11.8,5.35)
    {Output Distribution};

  \begin{scope}[on background layer]
    \draw[draw=black!45,fill=bankgray!55,line width=0.45pt]
      (2.1,-0.60) rectangle (9.5,4.3);
  \end{scope}
  \node[anchor=north west,font=\small\bfseries] at (2.25,4.65)
    {Thinking stage};

  \draw[flow] (input.north) -- (state.south);
  \draw[flow] (state.north) -- node[annotation,left] {$q_t$} (activate.south);
  \draw[flow] (activate.east) -- node[annotation,above] {$s_t$} (select.west);
  \draw[flow] (bank.south) -- (select.north);
  \draw[flow] (select.east) -- node[annotation,below=18pt] {$S_t,\alpha_t$} (assemble.west);
  \draw[flow] (assemble.south) -- node[annotation,left] {$\widetilde\kappa_t$} (think.north);
  \draw[flow] (state.east) -- (think.west);
  \draw[flow] (think.south) -- (8.2,0.05) -- (3.4,0.05) -- (state.south);
  \fill (3.4,0.05) circle (1.1pt);
  \node[annotation,anchor=north] at (5.8,-0.06)
    {up to $R$ steps; shared operator parameters};
  \draw[flow] (think.east) -- node[annotation,above=24pt,pos=0.9] {$X_{R_{\mathrm{exec}}}$} (generate.west);
  \node[annotation,anchor=north west] at (9.65,0.38) {at stopping};
  \draw[flow] (generate.north) --
    node[annotation,fill=white,inner sep=1pt] {Generation stage\\(optional)} (head.south);
  \draw[flow] (head.north) -- (output.south);
\end{tikzpicture}
\caption{Thinking followed by generation. Only the left stage recurs: its
Knowledge Activator and Knowledge Assembly configure a shared state-update
operator at every step. After $R_{\mathrm{exec}}$ updates, the final state enters $L$ ordinary
generation layers and an output head. The generation layers may retain
distinct pretrained parameters; when $L=0$, the final state goes directly to
the output head.}
\label{fig:think-generate}
\end{figure}

One configuration separates the recurrent state updates from an output stage
containing $L\ge0$ conventional blocks and a head, where $L$ is an integer.
Let $\Pi$ be the state-boundary map, $Z_\ell$ the complete state after $\ell$
output-stage blocks, $D_\ell^{\mathrm{gen}}$ the $\ell$th such block with its
parameters included, and $G_{\mathrm{head}}$ the final prediction head. Then
\begin{equation}
 \begin{aligned}
 Z_0&=\Pi(X_{R_{\mathrm{exec}}}),&
 Z_{\ell+1}&=D_{\ell+1}^{\mathrm{gen}}(Z_\ell;\mathcal C),
 \quad \ell=0,\ldots,L-1,\\
 Y&=G_{\mathrm{head}}(Z_L;\mathcal C).
 \end{aligned}
 \label{eq:thinking-generation}
\end{equation}
$\Pi$ preserves or adapts all state required by the output stage. $G_\varphi$
denotes that whole stage. $R$ controls recurrent computation; $L$ describes
output-stage depth, with $L=0$ permitting a direct head.
\Cref{fig:think-generate} makes the sharing boundary explicit.

\suppressfloats[t]

\subsubsection{Reusing the Pretrained Generation Tail}
The output stage can retain a pretrained tail and its original state-handling
rules. A fixed schedule for the preceding blocks then provides a
behavior-preserving initialization, as developed in \Cref{sec:migration}.
The resulting boundary separates recurrent state refinement from subsequent
output organization. Adaptation can focus on the recurrent pathway, the
knowledge-selection policy, and the state interface supplied to the retained
tail. This configuration preserves an established output pathway while
introducing a new organization of the preceding computation.

\subsection{Fine-Grained Knowledge Interfaces}
\subsubsection{Independent Banks for Local Computational Slots}
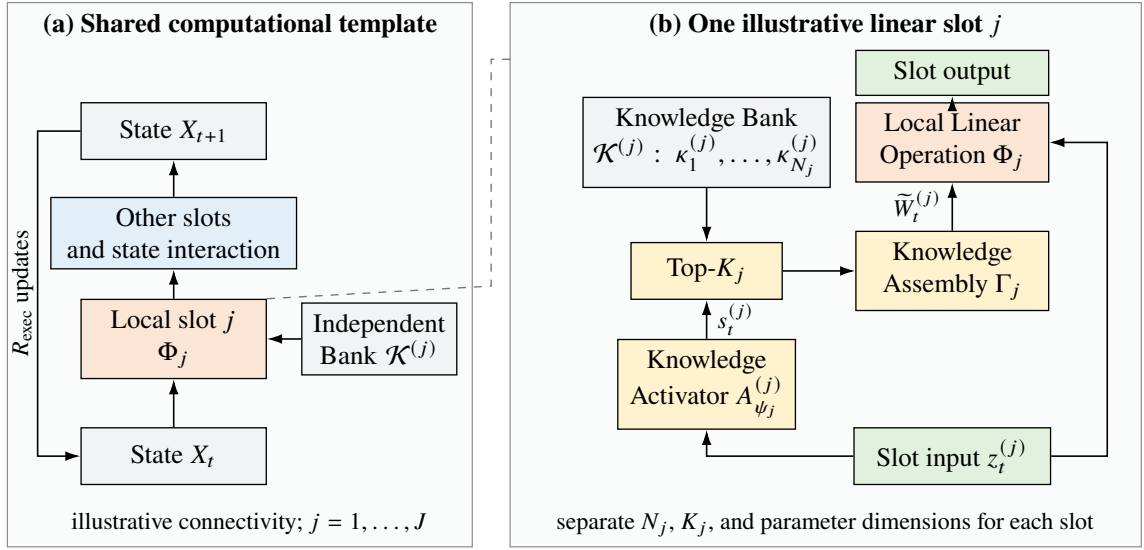
\begin{figure}[htbp]
\centering
\begin{tikzpicture}[
  x=1cm,y=1cm,font=\small,
  block/.style={draw=black,line width=0.5pt,rectangle,align=center,
    minimum height=0.75cm,inner sep=4pt},
  flow/.style={-{Latex[length=2mm,width=1.3mm]},line width=0.6pt},
  annotation/.style={font=\footnotesize,align=center}
]
  \begin{scope}[on background layer]
    \draw[draw=black!45,fill=bankgray!35,line width=0.45pt]
      (0,-0.55) rectangle (6.15,6.65);
    \draw[draw=black!45,fill=bankgray!35,line width=0.45pt]
      (6.65,-0.55) rectangle (15.0,6.65);
  \end{scope}
  \node[font=\small\bfseries] at (3.08,6.35)
    {(a) Shared computational template};
  \node[font=\small\bfseries] at (10.83,6.35)
    {(b) One illustrative linear slot $j$};

  \node[block,fill=bankgray,minimum width=2.45cm] (state) at (2.2,0.65)
    {State $X_t$};
  \node[block,fill=payloadpeach,minimum width=2.45cm] (slot) at (2.2,2.2)
    {Local slot $j$\\$\Phi_j$};
  \node[block,fill=bankgray,minimum width=2.05cm,text width=1.75cm] (localbank) at (4.92,2.2)
    {Independent\\Bank $\mathcal K^{(j)}$};
  \node[block,fill=computeblue,minimum width=3.25cm] (other) at (2.2,3.6)
    {Other slots\\and state interaction};
  \node[block,fill=bankgray,minimum width=2.45cm] (next) at (2.2,4.95)
    {State $X_{t+1}$};
  \draw[flow] (state.north) -- (slot.south);
  \draw[flow] (localbank.west) -- (slot.east);
  \draw[flow] (slot.north) -- (other.south);
  \draw[flow] (other.north) -- (next.south);
  \draw[flow] (next.west) -- (0.4,4.95) -- (0.4,0.65) -- (state.west);
  \node[annotation,rotate=90] at (0.2,2.8) {$R_{\mathrm{exec}}$ updates};
  \node[annotation,anchor=north] at (3.2,0.03)
    {illustrative connectivity; $j=1,\ldots,J$};

  \node[block,fill=outputgreen,minimum width=2.6cm] (input) at (12.5,0.65)
    {Slot input $z_t^{(j)}$};
  \node[block,fill=selectyellow,minimum width=2.35cm] (activate) at (9.25,1.6)
    {Knowledge\\Activator $A_{\psi_j}^{(j)}$};
  \node[block,fill=selectyellow,minimum width=2.0cm] (select) at (9.25,3.1)
    {Top-$K_j$};
  \node[block,fill=bankgray,minimum width=2.7cm] (bank) at (9.25,4.8)
    {Knowledge Bank\\
     $\mathcal K^{(j)}:\ \kappa^{(j)}_1,\ldots,\kappa^{(j)}_{N_j}$};
  \node[block,fill=selectyellow,minimum width=2.55cm] (assemble) at (12.5,3.1)
    {Knowledge\\Assembly $\Gamma_j$};
  \node[block,fill=payloadpeach,minimum width=2.55cm] (linear) at (12.5,4.8)
    {Local Linear\\Operation $\Phi_j$};
  \node[block,fill=outputgreen,minimum width=2.55cm,minimum height=0.55cm] (output)
    at (12.5,5.72) {Slot output};

  \draw[flow] (input.west) -- (9.25,0.65) -- (activate.south);
  \draw[flow] (activate.north) -- node[annotation,right] {$s_t^{(j)}$} (select.south);
  \draw[flow] (bank.south) -- (select.north);
  \draw[flow] (select.east) -- 
    (assemble.west);
  \draw[flow] (assemble.north) -- node[annotation,left] {$\widetilde W_t^{(j)}$} (linear.south);
  \draw[flow] (input.east) -- (14.55,0.65) -- (14.55,4.8) -- (linear.east);
  \draw[flow] (linear.north) -- (output.south);
  \node[annotation,anchor=north] at (10.82,0.03)
    {separate $N_j$, $K_j$, and parameter dimensions for each slot};
  \draw[dashed,line width=0.5pt,draw=black!55]
    (slot.north east) -- (6.4,2.9) -- (6.4,5.9) -- (6.65,5.9);
\end{tikzpicture}
\caption{Fine-grained Knowledge Banks. Each compatible local slot has its own
bank and activation policy. Panel (a) illustrates one slot inside a shared
state-update template; actual slots can be serial, parallel, or branching.
Panel (b) expands a linear slot: its input drives selection, the chosen
payloads configure the local matrix, and the same input is transformed by
the resulting operation. The dashed line is a correspondence guide.
Local slots are evaluated within each update of the outer recurrent template.}
\label{fig:slots}
\end{figure}

A template may expose an integer $J\ge1$ local interfaces, indexed by
$j\in\{1,\ldots,J\}$. Slot $j$ has a bank
$\mathcal K^{(j)}=\{\kappa_i^{(j)}\}_{i=1}^{N_j}$ with $N_j$ units,
an activator $A_{\psi_j}^{(j)}$ with parameters $\psi_j$, an assembly map
$\Gamma_j$, and a local operation $\Phi_j$.
At its invocation, the activator reads the slot input $z_t^{(j)}$ and returns
local scores $s_t^{(j)}$. It selects $K_j$ compatible units to configure
$\Phi_j$, where $N_j$ and $K_j$ are positive integers and $K_j$ does not
exceed the eligible count. Let $\mathcal I_t^{(j)}\subseteq\{1,\ldots,N_j\}$
be the eligible slot identifiers, so $s_t^{(j)}$ has
$|\mathcal I_t^{(j)}|$ real entries. For a linear slot, let
$d_{\mathrm{in}}^{(j)}$ and $d_{\mathrm{out}}^{(j)}$ denote its positive input
and output widths. Then $z_t^{(j)}\in\mathbb R^{d_{\mathrm{in}}^{(j)}}$ and
$\widetilde W_t^{(j)}\in\mathbb R^{d_{\mathrm{in}}^{(j)}\times d_{\mathrm{out}}^{(j)}}$,
with the input again represented as a row vector.
Unit identifiers are pairs $(j,i)$, with $i\in\{1,\ldots,N_j\}$.
The original dependency graph determines when the slot receives its input.
\Cref{fig:slots} illustrates this local invocation contract.

\suppressfloats[t]

\begin{samepage}
\subsubsection{Typed Templates and Heterogeneous Computation Graphs}
Different module types may use distinct shared operators, widths, banks, and
selection rules. Serial, parallel, and branched dependencies retain their
specified order. The outer recurrence reuses the whole typed graph, with
local banks supplying configurations at its slots. This organization
accommodates different slot counts, widths, and backbone topologies within
the same knowledge-invocation framework.
\par
\end{samepage}

\section{Static Realizations of Existing Language Models}
\label{sec:static}

A fixed-depth language model supplies a concrete reference path through the DeepThink interfaces. The construction below preserves that path before any dynamic policy is learned. It identifies a compatible static subfamily and provides the starting point for pretrained migration.

\subsection{The Realization Contract}
\subsubsection{Complete Configuration and Template Compatibility}

Write the pretrained backbone as transitions
\[
X^{(\ell)}=\mathcal F_{\ell,*}
  (X^{(\ell-1)};\mathcal C),\qquad \ell=1,\ldots,D,
\]
where \(X=(H,M)\) contains the complete execution state. Compatibility requires a shared template \(\Phi_\theta\) and configurations \(\kappa_\ell\) that realize these transitions. Configurations must retain every required parameter and layer-dependent rule, including normalization, biases, gates, and structural constants.

Here \(\mathcal F_{\ell,*}(X;\mathcal C)
:=F_\ell(X;\omega_\ell,\mathcal C)\) absorbs the original parameters into
the transition notation. We similarly use \(\mathcal E_*:=E_*\) and
\(\mathcal G_*:=G_*\) for the original initialization and output maps.

\subsubsection{Singleton Assembly and Preserved State Transitions}

For this construction, the execution interface accepts the complete unit
record as a configuration and consumes its type and payload. Singleton
assembly then returns the selected configuration unchanged:
\[
\Gamma(\{\kappa_\ell\},1)=\kappa_\ell.
\]
The input initialization, external context, output mapping, and state-transition rules are preserved. Cache entries remain associated with their original execution stages. The preserved cache identities and update schedules extend the realization to incremental generation.

\subsection{Exact Fixed-Schedule Realization}
\subsubsection{Proposition and Realization Conditions}

\begin{proposition}[Exact fixed-schedule realization]
\label[proposition]{prop:fixed}
Let a depth-\(D\) pretrained backbone satisfy
\[
\mathcal F_{\ell,*}(X;\mathcal C)
 =\Phi_\theta(X;\kappa_\ell,\mathcal C)
\]
on its reachable states, with complete configurations and preserved initialization, output, and runtime rules. Assume singleton assembly is the identity. The DeepThink instance
\[
N=D,\qquad R=D,\qquad K=1,\qquad
S_t=\{t+1\},\qquad \alpha_{t,t+1}=1,\quad t=0,\ldots,D-1
\]
has the same complete state after each corresponding transition and the same final output as the pretrained model.
This statement concerns mathematical transitions in real arithmetic. For stochastic execution it
holds pathwise under the matched random choices specified in
\Cref{app:realization}, and consequently preserves the output distribution.
\end{proposition}

The schedule retains the original layer order and gives every invocation its
original complete configuration. Initialization, state transitions, and the
output mapping therefore remain aligned with the pretrained execution.
\Cref{app:realization} specifies the deterministic and matched stochastic
execution contracts.

\subsubsection{Typed Realization of Heterogeneous Backbones}

\begin{corollary}[Typed realization]
\label[corollary]{cor:typed}
The same conclusion holds for a finite schedule of compatible typed templates when each scheduled template realizes its original transition and preserves the original interfaces and execution dependencies.
\end{corollary}

Let $\rho_\ell$ be the original type at stage $\ell$, and let
$\Phi^{(\rho)}$ denote the shared executor for type $\rho$, with its
parameters suppressed in this notation. The type schedule selects
$\Phi^{(\rho_\ell)}$ at stage $\ell$. Original adapters, branch dependencies, and retained intermediate states belong to the contract. The resulting schedule coordinates a family of shared typed operators while retaining the original backbone's interfaces and dependency graph.

\subsection{Interpretation of the Inclusion Relation}
\subsubsection{Representation-Level Containment}

Let $\mathcal F_{\mathrm{fixed}}^{\mathrm{compatible}}$ denote the input--output
function family of fixed-depth models satisfying the realization contract.
Let $\mathcal F_{\mathrm{DT}}$ denote the family admitted by the DeepThink
interfaces, including these original initialization and output maps. Then
\[
\mathcal F_{\mathrm{fixed}}^{\mathrm{compatible}}
 \subseteq \mathcal F_{\mathrm{DT}}.
\]
The inclusion represents the original parameter sequence as an addressable fixed schedule. DeepThink then exposes state-conditioned selection, repeated invocation, and compatible assembly as additional execution policies within the same framework.

\subsubsection{From Layer Positions to Addressable Configurations}
The construction transfers the original parameter information and execution
sequence into explicit knowledge interfaces. Initially, the shared component
can be the common execution template, with depth-specific numerical
parameters retained in the payloads. This behavior-preserving reference path
connects conventional layer stacks to dynamic knowledge execution. Shared
numerical cores and compressed payloads can then be introduced through the
migration steps in \Cref{sec:migration}.

\section{Migration from Pretrained Language Models}
\label{sec:migration}

Migration separates preservation of an existing function from learning new execution behavior. This allows a pretrained model to serve as the initial parameter source and reference model.

\subsection{A Behavior-Preserving Starting Point}
\subsubsection{Encapsulating Existing Parameters and Runtime Rules}

Register each original configuration as a knowledge unit while retaining input embeddings, position handling, causal rules, final processing, and output parameters. Preserve existing parameter ties where they matter to later updates. The shared template executes the original operations using the selected package and its associated state.

\subsubsection{Initializing the Original Execution Schedule}

Initialize the policy to select unit \(t+1\) at step \(t\), with weight one and \(R=D\). No learned selector is required to reproduce this schedule. Any additional iteration features or control inputs must initially be disabled or made behaviorally neutral. The result is an initialization that inherits the original function rather than relearning it.

\subsection{Shared-Base Reparameterization}
\subsubsection{Exact Reconstruction with Complete Deltas}

\begin{proposition}[Exact shared-base reparameterization]
\label[proposition]{prop:rebase}
For slot $j\in\{1,\ldots,J\}$ and depth $\ell\in\{1,\ldots,D\}$, let
$W_{\ell,*}^{(j)}\in\mathbb R^{d_{\mathrm{in}}^{(j)}\times d_{\mathrm{out}}^{(j)}}$
be the original matrix, where $d_{\mathrm{in}}^{(j)}$ and
$d_{\mathrm{out}}^{(j)}$ are that slot's input and output widths.
Choose a same-shaped shared base $W_{\mathrm{base}}^{(j)}$ and define
the complete difference $\Delta W_\ell^{(j)}$ by
\[
\Delta W_\ell^{(j)}
 =W_{\ell,*}^{(j)}-W_{\mathrm{base}}^{(j)}.
\]
Singleton selection with unit weight reconstructs
\[
W_{\mathrm{base}}^{(j)}+\Delta W_\ell^{(j)}
 =W_{\ell,*}^{(j)}
\]
exactly in real arithmetic. When all other parameters and runtime rules are retained, this rebasing preserves the fixed-schedule function.
\end{proposition}

A shared base establishes a common parameter reference, and complete deltas retain the original layer differences. The base can be selected to support subsequent adaptation and rank-constrained representations.

\subsubsection{Rank Conditions and Retained Approximation Residuals}

Choose a positive integer rank budget $r_j$ for slot $j$. A low-rank payload
uses $U_\ell^{(j)}\in\mathbb R^{d_{\mathrm{in}}^{(j)}\times r_j}$ and
$V_\ell^{(j)}\in\mathbb R^{r_j\times d_{\mathrm{out}}^{(j)}}$.
The residual $E_\ell^{(j)}$ has the same shape as $\Delta W_\ell^{(j)}$, giving
\[
\Delta W_\ell^{(j)}
 =U_\ell^{(j)}V_\ell^{(j)}+E_\ell^{(j)}.
\]
Keeping \(E_\ell^{(j)}\) preserves exact reconstruction. A representation without a residual is exact when its rank budget accommodates \(\operatorname{rank}(\Delta W_\ell^{(j)})\). Here $\operatorname{rank}$ is matrix rank. Low-rank adaptation supplies a
compact parameterization \citep{lora2021}; truncated singular-value
decomposition supplies a standard approximation \citep{eckart1936}.
\Cref{app:factoring} specifies the retained and discarded residual cases.

\subsection{Transitioning to Dynamic Execution}
\subsubsection{Learning Activation and Assembly Policies}

A trainable activator can first imitate the original schedule and then learn selection from the evolving state. Compatible assembly expands the available configurations, while task and retention objectives guide the resulting execution policies. Parameter interpolation, nonlinear expert-output mixtures, and reordered transitions each define a distinct policy family.

\subsubsection{Extending Knowledge Libraries and Recurrence Budgets}

New units can be introduced without renumbering existing ones, provided their types and assembly interfaces remain compatible. A larger library also changes the selector's task and potentially its routing cost. Reusing a unit at additional steps or increasing \(R\) changes the states and cache paths encountered. Adaptation learns how to use the expanded library and the newly reached runtime states.

New units can first be registered while the original fixed schedule remains active, then introduced into the learned selection policy during adaptation.

\subsection{A Staged Migration Procedure}
\subsubsection{Exact Encapsulation and Interface Refinement}

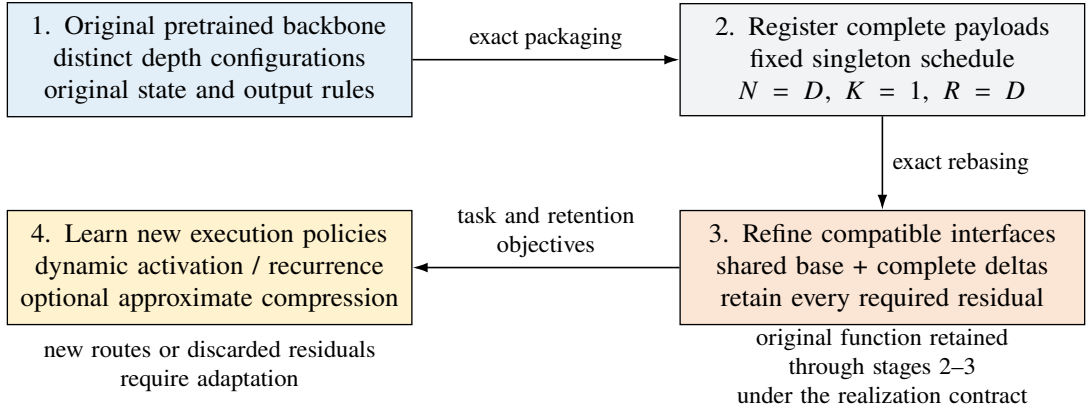
\begin{figure}[htbp]
\centering
\begin{tikzpicture}[
  x=1cm,y=1cm,font=\small,
  block/.style={draw=black,line width=0.5pt,rectangle,align=center,
    minimum height=1.4cm,inner sep=5pt},
  flow/.style={-{Latex[length=2mm,width=1.3mm]},line width=0.6pt},
  annotation/.style={font=\footnotesize,align=center}
]
  \node[block,fill=computeblue,text width=5.0cm] (original) at (3.05,4.25)
    {1. Original pretrained backbone\\
     distinct depth configurations\\original state and output rules};
  \node[block,fill=bankgray,text width=5.0cm] (registered) at (11.95,4.25)
    {2. Register complete payloads\\
     fixed singleton schedule\\$N=D,\ K=1,\ R=D$};
  \node[block,fill=payloadpeach,text width=5.0cm] (deltas) at (11.95,1.5)
    {3. Refine compatible interfaces\\
     shared base + complete deltas\\retain every required residual};
  \node[block,fill=selectyellow,text width=5.0cm] (dynamic) at (3.05,1.5)
    {4. Learn new execution policies\\
     dynamic activation / recurrence\\optional approximate compression};
  \draw[flow] (original.east) --
    node[annotation,above] {exact packaging} (registered.west);
  \draw[flow] (registered.south) --
    node[annotation,right] {exact rebasing} (deltas.north);
  \draw[flow] (deltas.west) --
    node[annotation,above,text width=2.9cm] {task and retention\\objectives} (dynamic.east);
  \node[annotation,text width=5.5cm] at (3.05,0.2)
    {new routes or discarded residuals\\require adaptation};
  \node[annotation,text width=5.5cm] at (11.95,0.2)
    {original function retained\\through stages 2--3\\under the realization contract};
\end{tikzpicture}
\caption{From pretrained execution to adaptive knowledge access. Complete
packaging and complete-delta rebasing retain the original fixed-schedule
function under the stated execution contract. Learning new routes or
discarding approximation residuals changes that reference computation.
Task and retention objectives guide adaptation after this boundary.
The diagram shows a full-backbone conversion; a retained output tail replaces
the full depth by the number of prefix blocks in the bank and recurrence counts.}
\label{fig:pretrained-migration}
\end{figure}

Migration proceeds through four stages: register the original configurations; expose typed slots or complete-delta representations; learn activation, assembly, and recurrence policies under task and retention objectives; and select compressed payload ranks for further adaptation. The first two stages retain the original schedule, providing a reference path
for the subsequent learned transformations. Exact interface refinement
retains complete payloads or complete deltas; dropping a residual belongs
to the compression stage. \Cref{fig:pretrained-migration} shows this boundary.

\suppressfloats[t]

\subsubsection{Adaptation, Retention, and Output-Stage Reuse}

Use the pretrained model as a teacher or retain its original schedule as supervision while learning new behavior. Recursive conversion with depth-specific low-rank modules already investigates pretrained initialization and parameter sharing \citep{rrt2024}; DeepThink places exact packaging and approximate compression within a common interface account.

The last $L$ original blocks can remain an output-organizing stage.
The preceding $D_{\mathrm{pre}}=D-L$ blocks then enter the Knowledge Bank,
where $D_{\mathrm{pre}}$ denotes the prefix depth.

\begin{corollary}[Exact reuse of the pretrained tail]
\label[corollary]{cor:tail}
Under the realization contract, choose an integer $0\le L<D$. Execute the
first $D_{\mathrm{pre}}=D-L$ complete configurations in their original
singleton order, with $N=R=D_{\mathrm{pre}}$ and $K=1$. Retaining the final
$L$ transitions and original output map as the generation stage preserves
the pretrained input--output function.
\end{corollary}

Let $\mathcal T_L$ denote this retained tail followed by the original output
map, with the original context and runtime rules. Its exact boundary is
\begin{equation}
X_{D_{\mathrm{pre}}}=X^{(D_{\mathrm{pre}})},\qquad
Y=\mathcal T_L(X_{D_{\mathrm{pre}}};\mathcal C)=Y_*.
\label{eq:tail-boundary}
\end{equation}
For $L=0$, $\mathcal T_0=\mathcal G_*$ and the prefix spans the entire
backbone. The retained tail supplies a ready output interface for adaptation,
starting from the original complete boundary state.

\section{Learning Objectives and Resource Considerations}
\label{sec:learning}

\subsection{Task and Retention Objectives}

\subsubsection{Language Modeling and Conditional Generation}
The execution interface determines how a state is produced; the task objective
determines which states and outputs are useful. Let $\mathcal D_{\mathrm{task}}$
be a distribution over input--target pairs $(x,y)$, with target sequence
$y=(y_1,\ldots,y_m)$ of length $m\ge1$. Let
$p_{\mathrm{DT}}(y\mid x;\mathcal C)$ denote its probability under DeepThink.
The negative-log-likelihood objective is
\begin{equation}
\mathcal L_{\mathrm{task}}
=\mathbb E_{(x,y)\sim\mathcal D_{\mathrm{task}}}
\left[-\log p_{\mathrm{DT}}(y\mid x;\mathcal C)\right].
\label{eq:task}
\end{equation}
For autoregressive generation, define the prefix
$y_{<p}=(y_1,\ldots,y_{p-1})$, with $y_{<1}$ empty. The likelihood factorizes as
\begin{equation}
p_{\mathrm{DT}}(y\mid x;\mathcal C)
=\prod_{p=1}^{m}p_{\mathrm{DT}}(y_p\mid x,y_{<p};\mathcal C).
\label{eq:autoregressive-likelihood}
\end{equation}
Every query used to predict position $p$ must observe the same legal
information boundary as its conditional probability. The input representation,
shared parameters, knowledge payloads, activator, and output stage may be
learned jointly or updated in selected groups. Training can initially
freeze pretrained payloads and output components while learning the activator,
then update selected parameter groups.

\subsubsection{Distribution Matching to the Pretrained Model}
A pretrained model can remain available as a teacher during migration,
following the distribution-matching principle of knowledge distillation
\citep{distill2015}. Let $\mathcal D_{\mathrm{retain}}$ be a distribution over
legal conditioning records $h$, each containing the known input and target
prefix. Let $p_*(\cdot\mid h;\mathcal C)$ and
$p_{\mathrm{DT}}(\cdot\mid h;\mathcal C)$ be the teacher and student next-token
distributions over a finite vocabulary $\mathcal V$.
One retention objective at unit temperature is
\begin{equation}
\mathcal L_{\mathrm{retain}}
=\mathbb E_{h\sim\mathcal D_{\mathrm{retain}}}
D_{\mathrm{KL}}\!\left(
p_*(\cdot\mid h;\mathcal C)
\,\|\,p_{\mathrm{DT}}(\cdot\mid h;\mathcal C)
\right).
\label{eq:retention}
\end{equation}
Here $D_{\mathrm{KL}}$ is Kullback--Leibler divergence. For distributions
$a$ and $b$ over $\mathcal V$, it is
$D_{\mathrm{KL}}(a\|b)=\sum_{v\in\mathcal V}a(v)\log[a(v)/b(v)]$,
where $v$ indexes vocabulary entries and the standard zero-probability
conventions apply. Under \Cref{prop:fixed}, the distributions coincide and
this term is zero in
the mathematical execution model. After the route, payload, or recurrence budget
changes, the same objective becomes an adaptation constraint. In practical
teacher forcing it can compare next-token distributions at matched prefixes.
The adaptation distribution determines which pretrained behaviors the
retention term emphasizes.

\subsection{Learning the New Interfaces}

\subsubsection{Discrete Selection and Singleton-Routing Gradients}
Hard Top-\(K\) selects a discrete support. Support selection and coefficient
weighting are separate optimization problems. For singleton activation,
normalization over the selected score gives
\begin{equation}
\alpha_{t,i}=1\quad\text{when }S_t=\{i\},
\qquad
\frac{\partial\alpha_{t,i}}{\partial s_{t,i}}=0.
\label{eq:singleton-gradient}
\end{equation}
Route imitation provides an initial supervised policy without differentiating
the discrete support. Let $i_t^*\in\mathcal I_t$ be the original scheduled
identifier, equal to $t+1$ in the complete-package realization. Let
$\mathcal D_{\mathrm{route}}$ sample original inputs and schedule positions
$(x,t)$, with scores evaluated on the corresponding legal execution states.
A full eligible-bank softmax distribution $\beta_t$ gives
\begin{equation}
\beta_{t,i}=\frac{\exp(s_{t,i})}{\sum_{u\in\mathcal I_t}\exp(s_{t,u})},
\qquad
\mathcal L_{\mathrm{route}}
=\mathbb E_{(x,t)\sim\mathcal D_{\mathrm{route}}}[-\log\beta_{t,i_t^*}].
\label{eq:route-imitation}
\end{equation}
Unlike selected-set normalization, this auxiliary distribution includes
unselected eligible units. Soft alternatives or gradient estimators provide
other approaches to training conditional selection \citep{bengio2013}.
The original schedule remains a reference policy.

\subsubsection{Load Balance and Functional-Specialization Objectives}
Usage constraints can discourage a small subset of units from receiving all
updates, a concern already addressed in sparse-expert training
\citep{shazeer2017,switch2021}. For an illustrative penalty, collect
$B_{\mathrm{inv}}\ge1$ routing decisions from one bank and index them by
$b=1,\ldots,B_{\mathrm{inv}}$. Use normalized nonnegative coefficients
$\alpha_{b,i}$, extended by zero outside each selected support. Define mean
usage $\bar\alpha_i$ and choose target proportions $\pi_i\ge0$ by
\begin{equation}
\bar\alpha_i=\frac{1}{B_{\mathrm{inv}}}\sum_{b=1}^{B_{\mathrm{inv}}}\alpha_{b,i},
\qquad
\Omega_{\mathrm{balance}}=\sum_{i=1}^{N}(\bar\alpha_i-\pi_i)^2,
\qquad \sum_{i=1}^{N}\pi_i=1.
\label{eq:balance}
\end{equation}
The target $\pi_i$ specifies the desired allocation of bank usage; this
squared penalty is one possible choice. With nonnegative coefficients
$\lambda_{\mathrm{retain}}$, $\lambda_{\mathrm{route}}$, and
$\lambda_{\mathrm{balance}}$, a combined learning objective is
\begin{equation}
\mathcal L=\mathcal L_{\mathrm{task}}
+\lambda_{\mathrm{retain}}\mathcal L_{\mathrm{retain}}
+\lambda_{\mathrm{route}}\mathcal L_{\mathrm{route}}
+\lambda_{\mathrm{balance}}\Omega_{\mathrm{balance}}.
\label{eq:combined-objective}
\end{equation}
The coefficients are hyperparameters chosen for the migration stage. Further objectives can target reuse of unit contributions
across contexts and coordination by the shared state-update operator.

\subsection{Parameters, Computation, and State}

\subsubsection{Stored Capacity Versus Active Parameter Use}
Parameter accounting distinguishes the bank from the subset used in one
update. Let $P_{\mathrm{stored}}$ count all distinct stored parameters.
The counts $P_{\mathrm{input}}$, $P_{\mathrm{shared}}$, $P_A$, $P_\Gamma$,
$P_G$, and $P_i$ cover initialization, the shared operator and bases,
query/activation/control, trainable assembly, the output stage, and unit $i$,
respectively. Assign tied tensors to one ownership group; a fixed interface
with no stored numerical parameters contributes zero. Then
\begin{equation}
P_{\mathrm{stored}}
=P_{\mathrm{input}}+P_{\mathrm{shared}}+P_A+P_\Gamma+P_G
+\sum_{i=1}^{N}P_i ,
\label{eq:storage}
\end{equation}
with tensors shared between groups counted once. A large \(N\) may coexist
with small \(K\), but all stored payloads, gradients, and optimizer state still
require an implementation strategy. For token-level routing, the union of
selected IDs across a sequence can exceed \(K\). Exact encapsulation retains
the original tensors; exact rebasing with full deltas can add storage.
Compression requires a further representation change, such as the explicitly
approximate case in \Cref{sec:migration}.

\subsubsection{Routing, Assembly, Recurrence, and Cache Ownership}
Let $C(x)$ be forward-execution cost for input $x$, measured in one consistent
unit such as floating-point operations. Let $C_E$ and $C_G$ be initialization
and output costs. At update $t$, let $C_A$ include query formation,
eligibility checks, activation, and stopping control; let $C_\Gamma$ and
$C_\Phi$ be assembly and state-transition costs. Then
\begin{equation}
C(x)=C_E(x)+\sum_{t=0}^{R_{\mathrm{exec}}-1}
\left[C_A(t,x)+C_\Gamma(t,x)+C_\Phi(t,x)\right]+C_G(x).
\label{eq:compute}
\end{equation}
Dense \(N\)-way scoring contributes \(N\)-dependent work alongside sparse
payload evaluation. Materializing merged matrices and evaluating
selected increments also have different arithmetic and data-movement costs.
Parameter sharing avoids additional parameter copies, while repeated
execution still produces work and, during training, intermediate activations.
Runtime-state ownership follows executed stages and their dependencies.
Caches track the inputs, routes, and position semantics of the computations
that produced them. \Cref{app:contracts} includes this ownership in the
execution interface.

\section{Research Landscape and Positioning}
\label{sec:landscape}

\DT{} connects several established research directions: conditional parameter access, recurrent state refinement, modular adaptation, and configurable output processing. This section organizes the primary literature available by 12 October 2026 around these mechanisms. The comparison focuses on execution interfaces, parameter organization, and the different design choices they expose.

\subsection{Conditional Capacity and Scalable Addressing}

\subsubsection{Sparse Expert Banks and Active Computation}

Sparse MoE models already distinguish stored parameter capacity from parameters activated for an input. The sparsely gated layer of \citet{shazeer2017} supplies a bank of expert subnetworks and a learned gate; Switch Transformers simplify routing to a single expert \citep{switch2021}. DeepSeekMoE combines fine-grained routed experts with shared experts \citep{deepseekmoe2024}. Consequently, a large bank with a small activation width is an established design principle. \DT{} adds a recurrent execution contract and an explicit assembly interface to this capacity-allocation perspective.

\subsubsection{Fine-Grained Experts and Product-Key Retrieval}

The addressing mechanism matters when the bank becomes very large. Product-key memory provides structured access to a large learned memory \citep{pkm2019}; Parameter Efficient Expert Retrieval (PEER) extends product-key retrieval to a bank of tiny experts \citep{peer2024}. These approaches are relevant to the proposed regime \(K\ll N\), because scoring every entry with a dense router can itself become expensive. Bank size and active expert count are therefore separate controls, while total execution cost still includes search, selection, and data movement.

\subsection{Recurrent Computation and Latent Refinement}

\subsubsection{Shared Depth and Adaptive Halting}

Universal Transformers apply shared transformations recurrently across depth and can use position-dependent halting \citep{ut2018}. Adaptive Computation Time supplies an earlier mechanism for learning how many recurrent updates to perform \citep{act2016}. These works establish that effective depth need not coincide with the number of distinct parameterized layers. They inform the shared transition \(\Phi_\theta\) and optional stopping interface of \DT{}. DeepThink builds on these recurrent mechanisms by making the selected parameter configuration an explicit input to each shared transition.

\subsubsection{Iterative Experts and Cross-Layer Expert Reuse}

Mixture-of-Experts Universal Transformers (MoEUT) is a particularly close comparator: it combines recurrent shared layer groups with routed feedforward experts and attention experts, including dynamically selected value and output projections \citep{moeut2024}. It already couples parameter reuse with changing expert access. Chain-of-Experts (CoE) instead repeatedly routes the evolving representation through experts within a layer, using iteration-specific routers and intermediate residual updates \citep{coe2025}. Mixture of Reused Experts (MoRE) shares feedforward expert pools across adjacent layers while retaining layer-specific attention, routers, and depth embeddings \citep{more2026}. Together, these works cover recurrent experts, iterative reselection, and cross-layer banks. DeepThink brings bank scope, payload types, assembly semantics, and pretrained execution into a common specification. The resulting interface vocabulary supports comparisons among these recurrent and cross-layer designs.

\subsubsection{Latent Feedback and Output-Stage Organization}

Recurrent-depth latent reasoning explicitly separates a prelude, a recurrent core, and a coda that produces the output distribution \citep{recurrent2025}. Thus, a recurrent computation stage followed by output organization already has a concrete precedent. Coconut feeds a final hidden state back as a subsequent input embedding during its latent mode \citep{coconut2024}. Its feedback advances sequence positions, whereas depth recurrence repeatedly updates a representation within the model's computational depth. Mixture-of-Recursions (MoR) routes token participation or recursion depth and studies recursion-aware key--value (KV) caching \citep{mor2025}. Its allocation primarily determines how much shared computation a token receives. \DT{} knowledge activation determines which compatible parameter content participates in an update; the two decisions can coexist. MoR also makes cache policy part of the execution design. Changing recurrence budgets can change which representations or cache entries are available to later tokens; the effect depends on the caching strategy. A budget controller therefore needs an explicit contract for state and attention behavior. This connection motivates preserving cache ownership explicitly when a pretrained execution path is changed.

\subsection{Modular Parameters and Runtime Assembly}

\subsubsection{Low-Rank Modules and Routed Adapter Libraries}

LoRA parameterizes adaptation through low-rank updates to pretrained matrices \citep{lora2021}. LoraHub composes previously trained task adapters using coefficients optimized from a few examples of a new task; composition is performed per task rather than per evolving hidden state \citep{lorahub2023}. MixLoRA introduces routed LoRA-based feedforward experts on a frozen pretrained model \citep{mixlora2024}, while MeteoRA learns gates for existing adapters at local linear interfaces \citep{meteora2024}. These methods directly inform low-rank knowledge libraries. Adapter reuse and input-conditioned module selection are already available mechanisms. Their granularity also matters: a task adapter may span many layers, whereas a local payload supplies one typed slot. A knowledge bank records both the payload's compatible interface and the level at which it is selected.

\subsubsection{Parameter Merging and Weight Generation}

SMEAR constructs a single expert by averaging all expert parameters under routing weights, then executes that merged expert \citep{smear2023}. It is a direct precedent for parameter-space assembly. HyperNetworks generate another network's weights from a learned network \citep{hyper2016}, offering a different source of conditional configurations. \DT{} may retrieve and assemble a finite bank without requiring arbitrary weight generation. The distinction between parameter mixing and mixing independently computed nonlinear expert outputs remains essential: they generally define different functions, even when their payloads have identical shapes.

\subsubsection{Pretrained Conversion and Depth-Specific Residuals}

Relaxed Recursive Transformers (RRT) initialize shared recursive blocks from pretrained models and use depth-specific LoRA modules to relax weight tying \citep{rrt2024}. Their residual-based initialization uses truncated singular value decomposition to approximate the difference between original and shared weights. This already establishes a migration route that reuses pretrained parameters. In \DT{}, complete residual payloads with the original singleton schedule supply an exact starting representation; reducing their rank can introduce approximation unless the discarded residual is retained. Moving from depth-indexed increments to state-conditioned selection is a further adaptation step. Task and retention objectives can guide the resulting state-conditioned combinations from this pretrained starting point.

\subsection{Knowledge Interfaces and Computational Templates}

\subsubsection{Parametric Memories, Knowledge Adapters, and Retrieval}

The interpretation of feedforward layers as key-value memories motivates examining addressable parameter content and the contributions of individual memory entries \citep{ffmemory2020}. K-Adapter supplies separately trained adapters for different kinds of infused knowledge while retaining a fixed pretrained backbone \citep{kadapter2020}. Retrieval-Augmented Generation combines parametric generation with retrieved nonparametric document content \citep{rag2020}. These are different knowledge interfaces. \DT{} focuses on callable parameter payloads and their place in recurrent execution; document retrieval and explicitly supervised knowledge adapters provide complementary reference points for defining what a knowledge unit is intended to represent.

\subsubsection{Fixed Templates and Conditional Representation}

The repeated attention and feedforward templates of the Transformer illustrate the separation of an operation graph from its numerical configuration \citep{vaswani2017}. The same interface perspective extends to other compatible language-model backbones.

Fixed-weight, input-programmable computation is established by prior work on in-context algorithm emulation. \citet{hualgorithm2026} construct fixed-weight softmax-attention models that approximately emulate prescribed finite libraries of attention-implementable algorithms through prompts. Their preprint appeared in August 2025. \citet{huuniversal2026} establish prompt-based universal approximation for Lipschitz continuous sequence-to-sequence functions on compact domains, together with one-pass and multi-pass composition. The emulator is fixed after specifying the function class, approximation accuracy, and composition-length budget. These guarantees extend beyond attention emulation.

Fixed Universal Transformers likewise simulate a specified class of Transformers using fixed internal parameters and suitable embeddings that encode target-model descriptions \citep{fixedut2026}. Across these works, the input specifies the target computation while the model weights remain fixed. They provide directly related theoretical foundations for conditional computation templates.

DeepThink develops this interface perspective through explicit selection and assembly of typed parameter payloads. Complete pretrained parameter packages support its exact Fixed-Schedule Realization under the specified execution contract. This construction connects fixed pretrained schedules to dynamically selected execution, while the cited prompt-programmable models encode target computations in their inputs.

\subsection{Comparative Synthesis}

\subsubsection{Mechanism-Level Comparison with Closest Architectures}

Tables~\ref{tab:landscape-execution} and~\ref{tab:landscape-assembly} compare execution controls, parameter composition, and pretrained starting points. Expert-content selection determines which transformations participate; token or budget routing determines where and how much computation occurs. Parameter-space assembly constructs an effective configuration, whereas function-space assembly combines computed outputs. Output stages organize the final state into predictions. Together, these distinctions identify the choices an implementation makes at each DeepThink interface.

\begin{table}[tbp]
\centering
\caption{Execution units, selection targets, and output organization in recurrent and modular architectures.}
\label{tab:landscape-execution}
\small
\begin{tabularx}{\linewidth}{@{}lYYY@{}}
\toprule
Method & Reuse or iteration unit & Selection target & Output organization \\
\midrule
MoEUT & Recurrent shared layer groups & Feedforward and attention experts & Prediction after recurrent groups \\
CoE & Expert iterations inside a layer & Experts chosen from updated states & Conventional outer model \\
MoRE & Shared feedforward pools across adjacent layers & Experts via layer-specific routers & Conventional outer layer stack \\
RRT & Repeated shared blocks & Depth-indexed LoRA modules & Original input/output components \\
\shortstack[l]{Recurrent\\depth} & Latent recurrent core & Number of core iterations & Prelude, core, and coda \\
Coconut & Hidden-state feedback across positions & Latent or language mode & Same language model switches modes \\
MoR & Shared stack with token-dependent depth & Token participation or recursion budget & Prediction after allocated depth \\
\DT{} & Shared typed state transitions & Compatible knowledge payloads & Explicit configurable output interface \\
\bottomrule
\end{tabularx}
\end{table}

\begin{table}[tbp]
\centering
\caption{Composition mechanisms and pretrained starting points. RRT denotes Relaxed Recursive Transformers.}
\label{tab:landscape-assembly}
\small
\begin{tabularx}{\linewidth}{@{}lYYY@{}}
\toprule
Method & Composition mechanism & Starting point & Connection to \DT{} \\
\midrule
LoraHub & Task-level adapter composition & Previously trained task adapters & Library reuse with task-level coefficients \\
MixLoRA & Routed LoRA-based expert outputs & Frozen pretrained dense model & Sparse adaptation at feedforward interfaces \\
MeteoRA & Gated local low-rank branches & Existing adapters and a base model & Input-conditioned slot-level payload access \\
SMEAR & Routing-weighted parameter average & Learned expert parameters & A merged expert before execution \\
HyperNetworks & Network-generated target weights & Generator and target trained together & Conditional configuration generation \\
RRT & Shared base plus depth-specific LoRA & Pretrained weights and residual initialization & Conversion with approximate low-rank deltas \\
MoEUT & Routed expert operations in shared groups & Trained shared expert architecture & Recurrent access to reusable expert banks \\
CoE & Successive mixtures of expert outputs & Trained iterative expert architecture & Updated-state selection over expert iterations \\
\DT{} & Typed parameter or function assembly & Complete pretrained payloads under fixed routing & Conditional exact realization before dynamic adaptation \\
\bottomrule
\end{tabularx}
\end{table}

\subsubsection{Established Mechanisms and the DeepThink Proposal}

The established ingredients include sparse capacity, shared recurrence, iterative expert choice, modular low-rank updates, parameter merging, and staged latent/output computation. DeepThink organizes these ingredients around addressable content, explicit assembly, and shared state updates. Fixed-Schedule Realization connects this organization to compatible pretrained computation; adaptation opens selection, composition, and recurrence to the evolving state. This perspective supports an integrated design process: choose the scope and type of the bank, define its assembly semantics, set the recurrent and output boundaries, and initialize those choices from an existing model.

\section{Design Implications and Open Questions}
\label{sec:implications}

\subsection{Potential Architectural Benefits}

\subsubsection{Independently Configurable Knowledge and Compute Controls}
DeepThink assigns separate controls to available payloads, active payloads,
recurrent updates, and output organization. Increasing \(N\) expands what
can be addressed; changing \(K\) changes how much participates in an update;
changing \(R\) changes the refinement budget; and \(L\) sets the depth of an
optional output tail. These choices expose a design space in which knowledge
capacity and computation can be adjusted separately. Their joint resource
accounting in \Cref{eq:storage,eq:compute} makes the resulting tradeoffs
concrete.

The Shared Thinking Operator reuses an execution structure across different
knowledge configurations. Successive updates can revisit a unit, combine
units differently, or move to another compatible bank as the state evolves.
This offers a direct way to express iterative refinement with changing
content, rather than assigning each configuration to a single depth position.

\subsubsection{Modular Updates and Pretrained Capability Reuse}
An addressable bank gives each update a concrete location and compatibility
interface. Units can be retained, replaced, extended, or frozen while common
input and output interfaces are preserved. This makes domain adaptation and
library growth expressible as local parameter and routing operations.
Activation traces expose how those operations are used across inputs and
iterations.

The constructions in \Cref{sec:static,sec:migration} supply a complete
pretrained path from which this organization can grow. Existing configurations
enter the bank, the original order initializes activation, and a retained
tail supplies a ready generation interface. A model can therefore begin with
pretrained capabilities and learn new execution policies through staged
adaptation. Interface refinement, parameter factoring, and dynamic activation
can be introduced as successive design choices.

\subsection{Functional Questions}

\subsubsection{Reusable Knowledge Units and Computational Organization}
The design intuition suggests units whose contributions remain useful across
different contexts and stages of computation. Their granularity is a central
choice: a complete configuration offers broad compatibility with pretrained
models, while local experts and low-rank modules offer finer control over
content selection. A shared operator then coordinates the selected
contributions through its state-update rules.

This raises questions about which unit types best support reuse, how
specialization emerges, and how knowledge composition interacts with
the evolving state. Replacing, updating, or withholding units provides a
concrete way to examine their roles. The alternative decompositions in
\Cref{app:factoring} also suggest studying how a chosen shared base and
payload constraints influence the functions learned by the bank and operator.

\subsubsection{Dynamic Execution, Retention, and Generalization}
A fixed schedule supplies an initial execution path. State-conditioned
activation opens a larger space of paths, including repeated invocation,
new combinations, and variable refinement budgets. Task and retention
objectives connect these choices to the original model's output behavior.
The retained generation stage provides a common boundary at which their
terminal states can be compared.

Promising questions concern activation across domains, useful stopping
criteria, composition at different slot types, and scalable addressing of
large banks. The interaction between recurrence and generation is equally
central: how should computation be allocated between refining a latent
state and organizing that state into language?

\subsection{Concluding Perspective}

\subsubsection{The DeepThink Perspective}
DeepThink views a language model as an evolving state that repeatedly calls
addressable knowledge configurations through shared computational interfaces.
Storage, selection, assembly, state updates, and output become explicit parts
of one framework. Conventional compatible layer stacks appear as fixed
schedules within this organization, providing a direct connection to
pretrained weights. Dynamic activation and assembly then turn configuration
access into a decision informed by the current state.

The resulting perspective links large knowledge libraries, sparse use,
recurrent refinement, and separate output organization. It provides a common
language for designing their interaction and a constructive starting point
for exploring that interaction in existing language models.

\subsubsection{Open Exploration}
This manuscript develops DeepThink through architectural intuition, module
definitions, and formal constructions. We have not conducted experiments.
The effectiveness of dynamic knowledge selection, the emergence of reusable
Knowledge Units, the allocation of recurrent and generation budgets, and the
retention of pretrained capabilities are open questions for exploration.
These questions invite concrete implementations and empirical study of how
addressable knowledge and shared thinking can work together.

\clearpage
\label{report:main-end}
\phantomsection
\addcontentsline{toc}{section}{References}
\renewcommand{\bibfont}{\small}
\setlength{\bibsep}{4pt}


\clearpage
\appendix
\label{report:appendix-start}

\section{Execution Contracts and Tail Reuse}
\label{app:realization}

\subsection{Complete Execution Contracts}
\subsubsection{Deterministic Execution and Matched Randomness}
Fix the input $x$, external context $\mathcal C$, and inference settings.
The original maps $\mathcal E_*$, $\mathcal F_{\ell,*}$, and $\mathcal G_*$
specify initialization, complete state transitions, and output.
The realization contract requires equal initialized states, compatible
complete configurations, identity singleton assembly, the original schedule,
and the original output map. Deterministic equality refers to mathematical
values in real arithmetic, including output probabilities.

For stochastic execution, let $\epsilon_\ell$ be the random input used at
stage $\ell$. Couple corresponding draws with the same conditional law and
require the transition contract
\begin{equation}
\mathcal F_{\ell,*}(X;\mathcal C,\epsilon_\ell)
=\Phi_\theta(X;\kappa_\ell,\mathcal C,\epsilon_\ell)
\label{eq:stochastic-contract}
\end{equation}
almost surely on reachable states. Randomized initialization and output use
the analogous matched contracts. Random-stream ownership and consumption
order are part of the execution specification; equal seeds alone do not
specify corresponding draws when call schedules differ.

\subsubsection{State Preservation and Layer-Dependent Configuration}
The runtime component $M$ contains every dynamic quantity that can influence
a later transition or output. This includes caches indexed by executed
stage, retained representations for skip paths, counters, and recurrent
working state. Layer-dependent parameters and structural constants belong
to complete payloads or preserved contextual rules. Positions and causal
masks remain in $\mathcal C$.

Implicit cache updates must appear in the declared state transition.
Corresponding executions use a common abstract representation of the
runtime objects, even when their physical storage layouts differ. A wrapper
counter is synchronized with the original stage index. Configuration
identity, cache association, and operation order are retained at each stage.

\subsection{Typed Execution Schedules}
\subsubsection{Compatible Executors and Graph Dependencies}
For the original type $\rho_\ell$ at stage $\ell$, the executor contract is
\begin{equation}
\mathcal F_{\ell,*}(X;\mathcal C)
=\Phi^{(\rho_\ell)}(X;\kappa_\ell,\mathcal C).
\label{eq:typed-contract}
\end{equation}
The fixed schedule records $(\rho_\ell,\kappa_\ell)$ for every original stage
in its original order. A branched graph retains all intermediate objects
needed by dependent branches. Different widths use compatible typed state
spaces or a workspace containing typed objects. Existing adapters between
these spaces retain their original semantics.

\subsection{Output-Stage Reuse}
\subsubsection{Prefix Boundary and Empty-Tail Convention}
For integer $0\le L<D$, the prefix depth is $D_{\mathrm{pre}}=D-L$.
\Cref{cor:tail} uses $N=R=D_{\mathrm{pre}}$, $K=1$, and the original prefix
schedule. The retained tail $\mathcal T_L$ receives the complete state
specified in \cref{eq:tail-boundary}, including any cache and skip-path
objects expected by the original tail. The boundary map $\Pi$ is the identity
on these corresponding abstract states.

For $L=0$, the output stage contains only the original output map:
$\mathcal T_0=\mathcal G_*$. The condition $L<D$ retains at least one
recurrent prefix update. Adaptation can subsequently change the prefix while
using the original boundary state as a retention target.

\section{Parameter Factoring and Approximation}
\label{app:factoring}

\subsection{Exact and Approximate Deltas}
\subsubsection{Full-Delta Reconstruction and Rank Conditions}
\Cref{prop:rebase} retains every complete difference
$\Delta W_\ell^{(j)}=W_{\ell,*}^{(j)}-W_{\mathrm{base}}^{(j)}$ together with
the original bias, normalization, and runtime configuration.
For a generic compatible matrix
$\Delta\in\mathbb R^{d_{\mathrm{in}}\times d_{\mathrm{out}}}$ and positive
integer rank budget $r$, let
$U\in\mathbb R^{d_{\mathrm{in}}\times r}$,
$V\in\mathbb R^{r\times d_{\mathrm{out}}}$, and
$E=\Delta-UV\in\mathbb R^{d_{\mathrm{in}}\times d_{\mathrm{out}}}$.
Residual-free reconstruction is available exactly when
$\operatorname{rank}(\Delta)\le r$, with zero padding if needed.
Retaining $E$ gives exact reconstruction for any chosen factors.

For truncated singular-value decomposition (SVD), let
$\sigma_a(\Delta)$ denote singular values in nonincreasing order, indexed by
positive integer $a$ and extended by zeros. The notation $\|\cdot\|_2$
denotes the Euclidean vector norm, and
$\opnorm{E}=\sup_{\|z\|_2=1}\|zE\|_2$ is the induced operator norm for
row-vector inputs. The discarded residual satisfies
\begin{equation}
\opnorm{E}=\sigma_{r+1}(\Delta).
\label{eq:svd-residual}
\end{equation}
Truncated SVD is a standard low-rank approximation
\citep{eckart1936}; exact residual-free reconstruction requires every omitted
singular value to be zero.

\subsubsection{Local Approximation Specifications}
For fixed row-vector input $z$, identical base and bias, and
$\Delta=UV+E$, the local output discrepancy is
\begin{equation}
\|z(W_{\mathrm{base}}+\Delta)-z(W_{\mathrm{base}}+UV)\|_2
=\|zE\|_2\le\|z\|_2\,\opnorm{E}.
\label{eq:local-residual-error}
\end{equation}
For a selected assembly, let $E_i$ be the same-shaped discarded residual of
unit $i$. Holding the input, support $S_t$, and coefficients $\alpha_{t,i}$
equal across the compared computations gives
\begin{equation}
\left\|z\sum_{i\in S_t}\alpha_{t,i}E_i\right\|_2
\le\|z\|_2\sum_{i\in S_t}|\alpha_{t,i}|\,\opnorm{E_i}.
\label{eq:assembled-residual-error}
\end{equation}
For nonnegative normalized coefficients, the right-hand side is at most
$\|z\|_2\max_{i\in S_t}\opnorm{E_i}$. These specifications concern one linear
interface with fixed routing, not a bound on the full recurrent model after
its states or routes change.

\subsection{Non-Unique Decompositions}
\subsubsection{Shared-Base Shift Invariance}
Let $B\in\mathbb R^{d_{\mathrm{in}}\times d_{\mathrm{out}}}$ be a compatible
shift matrix, and let primes mark a changed parameter convention:
\begin{equation}
W_{\mathrm{base}}'=W_{\mathrm{base}}+B,\qquad
\Delta W_i'=\Delta W_i-B.
\label{eq:base-shift}
\end{equation}
Singleton effective matrices are unchanged. Normalized weighted assembly
also retains its effective matrix:
\begin{equation}
W_{\mathrm{base}}'+\sum_{i\in S_t}\alpha_{t,i}\Delta W_i'
=W_{\mathrm{base}}+\sum_{i\in S_t}\alpha_{t,i}\Delta W_i,
\quad \sum_{i\in S_t}\alpha_{t,i}=1.
\label{eq:normalized-base-shift}
\end{equation}
Full-delta factoring admits multiple shared-base choices. A rank-constrained
library additionally requires every shifted delta to remain admissible.
Freezing a base or fixing a factor convention selects a parameter convention
for subsequent learning.

\subsubsection{Equivalent Low-Rank Factorizations}
For an invertible matrix $T\in\mathbb R^{r\times r}$, the factors
$U'=UT$ and $V'=T^{-1}V$ specify the same increment:
\begin{equation}
U'V'=(UT)(T^{-1}V)=UV.
\label{eq:factor-convention}
\end{equation}
The factors can therefore use different internal coordinates while retaining
the same effective matrix and rank budget. Constraints can select a
particular factor convention during optimization.

\section{Operational Specifications}
\label{app:contracts}

\subsection{Typed Module Interfaces}

\subsubsection{Knowledge Payload and Assembly Contracts}
A typed unit records an ID, type $\rho_i$, and payload $\vartheta_i$,
including structural constants required for execution. Its type specifies
dimensions, calling semantics, and a compatible template. Equal dimensions
alone do not make an attention projection, gate, and state transform
interchangeable.

Assembly receives compatible payloads, coefficients, and any required
ordering policy. It returns executable configuration $\widetilde\kappa_t$.
Fixed-schedule realization preserves the original singleton payload;
interpolation, additional units, or generated parameters require their
declared composition rule.

\subsubsection{State-Transition and Output Contracts}
An executor takes $X_t=(H_t,M_t)$, its configuration, and legal context
$\mathcal C$. It returns all mutable information required by later stages.
Type checking covers state layout and dependencies as well as tensor widths.
Updated metadata belongs to propagated state or a declared updated context.

The output stage consumes the terminal state through its boundary map.
Exact tail reuse preserves the original state interface. A sampled string
additionally requires a decoding policy; equal distributions give coupled
equal samples when that policy receives corresponding random draws.

\subsection{Forward Specifications}

\subsubsection{Fixed-Schedule Initialization}
\Cref{alg:fixed-realization} expresses the realization contract with
identifiers numbered from one. For type $\rho_i$, $\Phi_{\theta,\rho_i}$
denotes the compatible executor, preserving the original state transition.
Let $\tau_{\mathrm{dec}}$ denote the ordered decision trace;
$\langle\rangle$ is the empty sequence and $\gets$ denotes assignment.
The complete trace $\tau$ additionally records pre-update state snapshots.

\begingroup
\setlength{\intextsep}{6pt}
\begin{algorithm}[H]
\caption{Fixed-schedule realization}
\label{alg:fixed-realization}
\small
\begin{algorithmic}[1]
\Require $x$, legal context $\mathcal C$, and complete units
  $\kappa_1,\ldots,\kappa_D$.
\Ensure Original output $Y$ and decision trace $\tau_{\mathrm{dec}}$.
\Statex \textbf{Precondition:} Original unit order, compatible templates,
  and original runtime rules.
\State $X_0\gets\mathcal E_*(x;\mathcal C)$;
  $\tau_{\mathrm{dec}}\gets\langle\rangle$.
\For{$t=0,\ldots,D-1$}
  \State $i\gets t+1$;
    $\widetilde\kappa_t\gets\Gamma(\{\kappa_i\},(1))$
    \Comment{Identity singleton assembly}
  \State $X_{t+1}\gets\Phi_{\theta,\rho_i}
    (X_t,\widetilde\kappa_t;\mathcal C)$.
  \State Append identifier $i$ to $\tau_{\mathrm{dec}}$.
\EndFor
\State $Y\gets\mathcal G_*(X_D;\mathcal C)$;
  \Return $(Y,\tau_{\mathrm{dec}})$.
\end{algorithmic}
\end{algorithm}
\endgroup

For an output tail of length \(L\), the prefix schedule contains \(D-L\)
entries and the retained tail consumes the state that the original prefix
produced. The identity mapping is the appropriate boundary when no
representation change has been introduced.

\subsubsection{State-Dependent Dynamic Execution}
\Cref{alg:dynamic-execution} shares the same outer interface while changing
the selection rule. A typed controller limits admissible units, and a stop
rule defines the realized budget \(R_{\mathrm{exec}}\le R\).

\begingroup
\setlength{\intextsep}{6pt}
\begin{algorithm}[H]
\caption{State-dependent dynamic execution}
\label{alg:dynamic-execution}
\small
\begin{algorithmic}[1]
\Require $x$, legal context $\mathcal C$, bank $\mathcal K$,
  update budget $R\ge1$, and active width $K\ge1$.
\Ensure $Y$ and decision trace $\tau_{\mathrm{dec}}$.
\Statex \textbf{Precondition:} $K\le|\mathcal I_t|$ at every executed update.
\State $X_0\gets E_\xi(x;\mathcal C)$;
  $\tau_{\mathrm{dec}}\gets\langle\rangle$; $R_{\mathrm{exec}}\gets0$.
\For{$t=0,\ldots,R-1$}
  \State $q_t\gets Q(X_t,\mathcal C,t)$;
    determine compatible identifiers $\mathcal I_t$.
  \State $s_t\gets A_\psi(q_t,X_t,\mathcal C,t;\mathcal I_t)$;
    $S_t\gets\operatorname{TopK}_K(s_t;\mathcal I_t)$.
  \State Compute coefficients $\alpha_t$ on $S_t$;
    $\widetilde\kappa_t\gets\Gamma(\mathcal K|_{S_t},\alpha_t)$.
  \State $X_{t+1}\gets\Phi_\theta(X_t,\widetilde\kappa_t;\mathcal C)$;
    $R_{\mathrm{exec}}\gets t+1$.
  \State Append $(S_t,\alpha_t)$ to $\tau_{\mathrm{dec}}$.
  \If{the declared stopping criterion holds for $X_{t+1}$}
    \State Terminate the loop.
  \EndIf
\EndFor
\State $Y\gets G_\varphi(X_{R_{\mathrm{exec}}};\mathcal C)$;
  \Return $(Y,\tau_{\mathrm{dec}})$.
\end{algorithmic}
\end{algorithm}
\endgroup

Token-local and slot-local selection are handled by the called interfaces. The static schedule supplies a reference policy alongside learned dynamic execution.

\subsection{Routing and Runtime State}

\subsubsection{Causal Queries and Order-Sensitive Assembly}
At target position \(p\), a query is measurable with respect to the known
prefix and legal external conditions. It cannot pool unknown future target
states and then broadcast that choice to earlier positions. This requirement
applies to activators and controllers as well as to the model's attention
mask. Input conditioning that is already known in full is a separate part
of the execution context.

A support set specifies a weighted matrix sum. Sequential composition uses
an ordered list. Let $f_i$ be the callable transformation supplied by
compatible unit $i$, and let $\circ$ denote function composition. For two
selected identifiers $i_1$ and $i_2$, $f_{i_2}\circ f_{i_1}$ and
$f_{i_1}\circ f_{i_2}$ generally differ. The assembly interface
records the intended operation. Parameter mixing followed by a nonlinearity
and averaging nonlinear expert outputs likewise define different functions.

\subsubsection{Cache Ownership under Changed Execution Paths}
Runtime caches are owned by the executed computation, its parameter
configuration, and its dependencies. Different stages that share parameters may
receive different representations and produce different keys, values or
recurrent state. Exact initialization therefore retains the original cache
identity and update schedule.

Changed units, coefficients, order, or budgets require a dependency check.
Uncombined projections and route-dependent mixtures have different cache
validity conditions. Reuse tracks input, route, position, and assembly
dependencies.


\begin{thebibliography}{32}

\bibitem[Backus(1978)]{backus1978}
John Backus.
\newblock Can Programming Be Liberated from the von Neumann Style? A Functional Style and Its Algebra of Programs.
\newblock \emph{Communications of the ACM}, 21(8):613--641, 1978.
\newblock \url{https://doi.org/10.1145/359576.359579}.

\bibitem[Bae et~al.(2025a)]{rrt2024}
Sangmin Bae, Adam Fisch, Hrayr Harutyunyan, Ziwei Ji, Seungyeon Kim, and Tal Schuster.
\newblock Relaxed Recursive Transformers: Effective Parameter Sharing with Layer-wise LoRA.
\newblock \emph{International Conference on Learning Representations (ICLR)}, 2025a.
\newblock \url{https://arxiv.org/abs/2410.20672}.

\bibitem[Bae et~al.(2025b)]{mor2025}
Sangmin Bae, Yujin Kim, Reza Bayat, Sungnyun Kim, Jiyoun Ha, Tal Schuster, Adam Fisch, Hrayr Harutyunyan, Ziwei Ji, Aaron Courville, and Se-Young Yun.
\newblock Mixture-of-Recursions: Learning Dynamic Recursive Depths for Adaptive Token-Level Computation.
\newblock \emph{arXiv preprint arXiv:2507.10524}, 2025b.
\newblock \url{https://arxiv.org/abs/2507.10524}.

\bibitem[Bengio et~al.(2013)]{bengio2013}
Yoshua Bengio, Nicholas L\'{e}onard, and Aaron Courville.
\newblock Estimating or Propagating Gradients Through Stochastic Neurons for Conditional Computation.
\newblock \emph{arXiv preprint arXiv:1308.3432}, 2013.
\newblock \url{https://arxiv.org/abs/1308.3432}.

\bibitem[Csord\'{a}s et~al.(2024)]{moeut2024}
R\'{o}bert Csord\'{a}s, Kazuki Irie, J\"{u}rgen Schmidhuber, Christopher Potts, and Christopher D. Manning.
\newblock MoEUT: Mixture-of-Experts Universal Transformers.
\newblock \emph{arXiv preprint arXiv:2405.16039}, 2024.
\newblock \url{https://arxiv.org/abs/2405.16039}.

\bibitem[Dai et~al.(2024)]{deepseekmoe2024}
Damai Dai, Chengqi Deng, Chenggang Zhao, R.X. Xu, Huazuo Gao, Deli Chen, Jiashi Li, Wangding Zeng, Xingkai Yu, Y. Wu, Zhenda Xie, Y.K. Li, Panpan Huang, Fuli Luo, Chong Ruan, Zhifang Sui, and Wenfeng Liang.
\newblock DeepSeekMoE: Towards Ultimate Expert Specialization in Mixture-of-Experts Language Models.
\newblock \emph{arXiv preprint arXiv:2401.06066}, 2024.
\newblock \url{https://arxiv.org/abs/2401.06066}.

\bibitem[Dehghani et~al.(2019)]{ut2018}
Mostafa Dehghani, Stephan Gouws, Oriol Vinyals, Jakob Uszkoreit, and \L{}ukasz Kaiser.
\newblock Universal Transformers.
\newblock \emph{International Conference on Learning Representations (ICLR)}, 2019.
\newblock \url{https://arxiv.org/abs/1807.03819}.

\bibitem[Eckart and Young(1936)]{eckart1936}
Carl Eckart and Gale Young.
\newblock The Approximation of One Matrix by Another of Lower Rank.
\newblock \emph{Psychometrika}, 1(3):211--218, 1936.
\newblock \url{https://doi.org/10.1007/BF02288367}.

\bibitem[Fedus et~al.(2022)]{switch2021}
William Fedus, Barret Zoph, and Noam Shazeer.
\newblock Switch Transformers: Scaling to Trillion Parameter Models with Simple and Efficient Sparsity.
\newblock \emph{Journal of Machine Learning Research, 23(120):1--39}, 2022.
\newblock \url{https://www.jmlr.org/papers/v23/21-0998.html}.

\bibitem[Geiping et~al.(2025)]{recurrent2025}
Jonas Geiping, Sean McLeish, Neel Jain, John Kirchenbauer, Siddharth Singh, Brian R. Bartoldson, Bhavya Kailkhura, Abhinav Bhatele, and Tom Goldstein.
\newblock Scaling up Test-Time Compute with Latent Reasoning: A Recurrent Depth Approach.
\newblock \emph{arXiv preprint arXiv:2502.05171}, 2025.
\newblock \url{https://arxiv.org/abs/2502.05171}.

\bibitem[Geva et~al.(2021)]{ffmemory2020}
Mor Geva, Roei Schuster, Jonathan Berant, and Omer Levy.
\newblock Transformer Feed-Forward Layers Are Key-Value Memories.
\newblock \emph{Proceedings of EMNLP, pages 5484--5495}, 2021.
\newblock \url{https://aclanthology.org/2021.emnlp-main.446/}.

\bibitem[Graves(2016)]{act2016}
Alex Graves.
\newblock Adaptive Computation Time for Recurrent Neural Networks.
\newblock \emph{arXiv preprint arXiv:1603.08983}, 2016.
\newblock \url{https://arxiv.org/abs/1603.08983}.

\bibitem[Ha et~al.(2016)]{hyper2016}
David Ha, Andrew Dai, and Quoc V. Le.
\newblock HyperNetworks.
\newblock \emph{arXiv preprint arXiv:1609.09106}, 2016.
\newblock \url{https://arxiv.org/abs/1609.09106}.

\bibitem[Hao et~al.(2024)]{coconut2024}
Shibo Hao, Sainbayar Sukhbaatar, DiJia Su, Xian Li, Zhiting Hu, Jason Weston, and Yuandong Tian.
\newblock Training Large Language Models to Reason in a Continuous Latent Space.
\newblock \emph{arXiv preprint arXiv:2412.06769}, 2024.
\newblock \url{https://arxiv.org/abs/2412.06769}.

\bibitem[He(2024)]{peer2024}
Xu Owen He.
\newblock Mixture of A Million Experts.
\newblock \emph{arXiv preprint arXiv:2407.04153}, 2024.
\newblock \url{https://arxiv.org/abs/2407.04153}.

\bibitem[Hinton et~al.(2015)]{distill2015}
Geoffrey Hinton, Oriol Vinyals, and Jeff Dean.
\newblock Distilling the Knowledge in a Neural Network.
\newblock \emph{arXiv preprint arXiv:1503.02531}, 2015.
\newblock \url{https://arxiv.org/abs/1503.02531}.

\bibitem[Hu et~al.(2022)]{lora2021}
Edward J. Hu, Yelong Shen, Phillip Wallis, Zeyuan Allen-Zhu, Yuanzhi Li, Shean Wang, Lu Wang, and Weizhu Chen.
\newblock LoRA: Low-Rank Adaptation of Large Language Models.
\newblock \emph{International Conference on Learning Representations (ICLR)}, 2022.
\newblock \url{https://openreview.net/forum?id=nZeVKeeFYf9}.

\bibitem[Hu et~al.(2026a)]{hualgorithm2026}
Jerry Yao-Chieh Hu, Hude Liu, Jennifer Yuntong Zhang, and Han Liu.
\newblock In-Context Algorithm Emulation in Fixed-Weight Transformers.
\newblock \emph{International Conference on Learning Representations (ICLR)}, 2026a.
\newblock \url{https://openreview.net/forum?id=BC7YLA0zJ0}.
\newblock Preprint first posted August 2025: \url{https://arxiv.org/abs/2508.17550}.

\bibitem[Hu et~al.(2026b)]{huuniversal2026}
Jerry Yao-Chieh Hu, Hong-Yu Chen, Po-Chiao Lin, Maojiang Su, and Han Liu.
\newblock In-Context Universal Approximation, Compositional Generalization, and Algorithm Emulation.
\newblock \emph{Proceedings of the 43rd International Conference on Machine Learning (ICML), Proceedings of Machine Learning Research}, 306:44705--44798, 2026b.
\newblock \url{https://proceedings.mlr.press/v306/hu26h.html}.

\bibitem[Huang et~al.(2024)]{lorahub2023}
Chengsong Huang, Qian Liu, Bill Yuchen Lin, Tianyu Pang, Chao Du, and Min Lin.
\newblock LoraHub: Efficient Cross-Task Generalization via Dynamic LoRA Composition.
\newblock \emph{Conference on Language Modeling (COLM)}, 2024.
\newblock \url{https://openreview.net/forum?id=TrloAXEJ2B}.

\bibitem[Lample et~al.(2019)]{pkm2019}
Guillaume Lample, Alexandre Sablayrolles, Marc'Aurelio Ranzato, Ludovic Denoyer, and Herv\'{e} J\'{e}gou.
\newblock Large Memory Layers with Product Keys.
\newblock \emph{arXiv preprint arXiv:1907.05242}, 2019.
\newblock \url{https://arxiv.org/abs/1907.05242}.

\bibitem[Lewis et~al.(2020)]{rag2020}
Patrick Lewis, Ethan Perez, Aleksandra Piktus, Fabio Petroni, Vladimir Karpukhin, Naman Goyal, Heinrich K\"{u}ttler, Mike Lewis, Wen-tau Yih, Tim Rockt\"{a}schel, Sebastian Riedel, and Douwe Kiela.
\newblock Retrieval-Augmented Generation for Knowledge-Intensive NLP Tasks.
\newblock \emph{arXiv preprint arXiv:2005.11401}, 2020.
\newblock \url{https://arxiv.org/abs/2005.11401}.

\bibitem[Li et~al.(2024)]{mixlora2024}
Dengchun Li, Yingzi Ma, Naizheng Wang, Zhengmao Ye, Zhiyuan Cheng, Yinghao Tang, Yan Zhang, Lei Duan, Jie Zuo, Cal Yang, and Mingjie Tang.
\newblock MixLoRA: Enhancing Large Language Models Fine-Tuning with LoRA-based Mixture of Experts.
\newblock \emph{arXiv preprint arXiv:2404.15159}, 2024.
\newblock \url{https://arxiv.org/abs/2404.15159}.

\bibitem[Liu et~al.(2026)]{fixedut2026}
Jingwen Liu, Alexandr Andoni, and Daniel Hsu.
\newblock Fixed Universal Transformers.
\newblock \emph{arXiv preprint arXiv:2605.31423}, 2026.
\newblock \url{https://arxiv.org/abs/2605.31423}.

\bibitem[Muqeeth et~al.(2023)]{smear2023}
Mohammed Muqeeth, Haokun Liu, and Colin Raffel.
\newblock Soft Merging of Experts with Adaptive Routing.
\newblock \emph{arXiv preprint arXiv:2306.03745}, 2023.
\newblock \url{https://arxiv.org/abs/2306.03745}.

\bibitem[Qiu et~al.(2026)]{more2026}
Eric S. Qiu, Utku Umur Acikalin, Justin Lovelace, Christian Belardi, Arjun B. Mulchandani, Carla P. Gomes, and Kilian Q. Weinberger.
\newblock MoRE: Mixture of Reused Experts.
\newblock \emph{arXiv preprint arXiv:2609.18176}, 2026.
\newblock \url{https://arxiv.org/abs/2609.18176}.

\bibitem[Shazeer et~al.(2017)]{shazeer2017}
Noam Shazeer, Azalia Mirhoseini, Krzysztof Maziarz, Andy Davis, Quoc Le, Geoffrey Hinton, and Jeff Dean.
\newblock Outrageously Large Neural Networks: The Sparsely-Gated Mixture-of-Experts Layer.
\newblock \emph{arXiv preprint arXiv:1701.06538}, 2017.
\newblock \url{https://arxiv.org/abs/1701.06538}.

\bibitem[Vaswani et~al.(2017)]{vaswani2017}
Ashish Vaswani, Noam Shazeer, Niki Parmar, Jakob Uszkoreit, Llion Jones, Aidan N. Gomez, Lukasz Kaiser, and Illia Polosukhin.
\newblock Attention Is All You Need.
\newblock \emph{arXiv preprint arXiv:1706.03762}, 2017.
\newblock \url{https://arxiv.org/abs/1706.03762}.

\bibitem[von Neumann(1945)]{vonneumann1945}
John von Neumann.
\newblock First Draft of a Report on the EDVAC.
\newblock Moore School of Electrical Engineering, University of Pennsylvania, 1945.
\newblock Digitized original: \url{https://doi.org/10.5479/sil.538961.39088011475779}.

\bibitem[Wang et~al.(2020)]{kadapter2020}
Ruize Wang, Duyu Tang, Nan Duan, Zhongyu Wei, Xuanjing Huang, Jianshu Ji, Guihong Cao, Daxin Jiang, and Ming Zhou.
\newblock K-Adapter: Infusing Knowledge into Pre-Trained Models with Adapters.
\newblock \emph{arXiv preprint arXiv:2002.01808}, 2020.
\newblock \url{https://arxiv.org/abs/2002.01808}.

\bibitem[Wang et~al.(2025)]{coe2025}
Zihan Wang, Rui Pan, Jiarui Yao, Robert Csordas, Linjie Li, Lu Yin, Jiajun Wu, Tong Zhang, Manling Li, and Shiwei Liu.
\newblock Chain-of-Experts: Unlocking the Communication Power of Mixture-of-Experts Models.
\newblock \emph{arXiv preprint arXiv:2506.18945}, 2025.
\newblock \url{https://arxiv.org/abs/2506.18945}.

\bibitem[Xu et~al.(2024)]{meteora2024}
Jingwei Xu, Junyu Lai, and Yunpeng Huang.
\newblock MeteoRA: Multiple-tasks Embedded LoRA for Large Language Models.
\newblock \emph{arXiv preprint arXiv:2405.13053}, 2024.
\newblock \url{https://arxiv.org/abs/2405.13053}.

\end{thebibliography}
\end{document}